\documentclass[letterpaper,journal]{IEEEtran}

\usepackage{amsmath,amssymb,amsfonts}

\usepackage{algorithm}
\usepackage{bm}

\usepackage{array}
\usepackage[caption=false,font=normalsize,labelfont=sf,textfont=sf]{subfig}
\usepackage{textcomp}
\usepackage{stfloats}
\usepackage{url}
\usepackage{verbatim}
\usepackage{tikz}      
\usepackage{booktabs}
\usetikzlibrary{calc,patterns,decorations.pathmorphing,decorations.markings,circuits}
\usepackage{cite}
\usepackage{xcolor}
\usepackage{centernot}
\usepackage{fix-cm}
\usetikzlibrary{positioning,calc,arrows.meta}

\usepackage[ruled,vlined,algo2e]{algorithm2e}

\newtheorem{assumption}{Assumption}
\newtheorem{lem}{Lemma}
\newtheorem{rem}{Remark}
\newtheorem{thm}{Theorem}

\newtheorem{definition}{Definition}
\newtheorem{problem}{Problem}
\def\BibTeX{{\rm B\kern-.05em{\sc i\kern-.025em b}\kern-.08em
    T\kern-.1667em\lower.7ex\hbox{E}\kern-.125emX}}
\usepackage{algorithm}
\usepackage{algpseudocode}

\begin{document}
\title{Robust Model Predictive Control Design for Autonomous Vehicles with Perception-based Observers}
\author{Nariman~Niknejad,~\IEEEmembership{Graduate Student Member,~IEEE},  Gokul S. Sankar, Bahare Kiumarsi,~\IEEEmembership{Member,~IEEE}, and Hamidreza~Modares,~\IEEEmembership{Senior Member,~IEEE} ~
\thanks{This work is supported by the Ford Motor Company–Michigan State University Alliance. \textit{Corresponding author: Hamidreza Modares.}}
\thanks{N. Niknejad and H. Modares are with the Department of Mechanical Engineering, Michigan State University, East Lansing, MI, 48863, USA. (e-mails:niknejad@msu.edu, modaresh@msu.edu).}
\thanks{B. Kiumarsi is with the Department of Electrical and Computer Engineering, Michigan State University, East Lansing, MI 48824, USA. (e-mail:kiumarsi@msu.edu).}
\thanks{G. S. Sankar is an independent researcher in MI, USA. (email: gokul.sivasankar@gmail.com).}%
}
\markboth{Journal of \LaTeX\ Class Files,~Vol.~18, No.~9, September~2020}%
{How to Use the IEEEtran \LaTeX \ Templates}

\maketitle

\begin{abstract}
   This paper presents a robust model predictive control (MPC) framework that explicitly addresses the non-Gaussian noise inherent in deep learning–based perception modules used for state estimation. Recognizing that accurate uncertainty quantification of the perception module is essential for safe feedback control, our approach departs from the conventional assumption of zero-mean noise quantification of the perception error. Instead, it employs set-based state estimation with constrained zonotopes to capture biased, heavy-tailed uncertainties while maintaining bounded estimation errors. To improve computational efficiency, the robust MPC is reformulated as a linear program (LP), using a Minkowski–Lyapunov-based cost function with an added slack variable to prevent degenerate solutions. Closed-loop stability is ensured through Minkowski–Lyapunov inequalities and contractive zonotopic invariant sets. The largest stabilizing terminal set and its corresponding feedback gain are then derived via an ellipsoidal approximation of the zonotopes. The proposed framework is validated through both simulations and hardware experiments on an omnidirectional mobile robot along with a camera and a convolutional neural network–based perception module implemented within a ROS2 framework. The results demonstrate that the perception-aware MPC provides stable and accurate control performance under heavy-tailed noise conditions, significantly outperforming traditional Gaussian-noise-based designs in terms of both state estimation error bounding and overall control performance.
\end{abstract}


\section{Introduction}


\IEEEPARstart{R}{ecently}, there has been a surge of interest in integrating high-dimensional perception modules (e.g., cameras, LiDAR) with advanced feedback controllers \cite{lin2024vision, chang2022active, zheng2023robust}. In autonomous driving systems, the perception module is essential not only for sensing the surrounding environment but also for rapidly adapting to dynamic environmental changes to ensure safety and stability \cite{he2024vision}. 

Robustness of the perception module against adversarial attacks and model uncertainties in deep learning-based image processing has been extensively studied in the literature \cite{abhulimhen2025multi}. When integrated with control strategies, two primary perception-based control approaches have emerged to address the compounded challenges of sensor noise and control performance. The first strategy directly converts raw sensor data into control actions using end-to-end frameworks \cite{li2015vision, rausch2017learning, huang2025leader, chen2024end}, while the second approach relies on deep learning models to extract system states for use in traditional feedback control loops \cite{dean2021certainty, dean2020robust, al2020accuracy, li2024safe}. A hybrid approach is also leveraged in \cite{nagariya2025learning}. Despite the success of conventional neural networks (CNNs) in computer vision tasks, their application in state estimation remains limited, lacking robust safety guarantees \cite{rahimi2022robust}. A challenge is that conventional uncertainty models for the perception modules that use CNNs and other neural networks assume Gaussian, zero-mean white noise uncertainties often fail to capture the complex and structured noise characteristics encountered in real-world applications \cite{al2020accuracy, wei2023multi,rahimi2022robust, maharmeh2025comprehensive}. This poses multiple challenges for their deployment in safety-critical scenarios \cite{bai2024survey}, and has motivated the adoption of set-membership approaches that model perception map uncertainties as unknown but bounded sets \cite {schweppe2003recursive, combastel2015zonotopes}.

To design a robust controller with safety and performance guarantees while considering uncertainties (not necessarily perception uncertainties), tube-based model predictive control (MPC) has emerged as an effective approach to mitigate process and measurement noise \cite{qie2022improved}, by taking advantage of set invariance principles \cite{greiff2024invariant}. While traditional MPC is widely recognized for its effective performance optimization and constraint handling, \cite{liu2023tube}, the tube-based approach further enhances system reliability by ensuring robust constraint satisfaction. This is achieved by deciding on nominal control actions to ensure that all predicted nominal trajectories remain within precomputed disturbance-induced invariant sets (i.e., tubes) \cite{ limon2010robust,fontes2017rigid}. Although several explicit set-theoretic methods have been proposed for constructing these tubes and determining their gains \cite{rakovic2023implicit, diaconescu2024elastic}, a computationally efficient formulation of tube-based MPC that computes both the tubes and nominal control actions is still lacking in the literature, despite their critical importance for real-time, onboard implementation in autonomous systems. MPC typically uses a quadratic cost function, resulting in a quadratic program that is computationally intensive. This choice is motivated by the shortcomings of linear-programming–based MPC, which often exhibits idle and deadbeat behavior due to the nonsmooth 1-norm cost function \cite{saffer2004analysis}. While set‐theoretic techniques \cite{ raghuraman2022set} have proven effective for computing robust gains under diverse noise profiles, their use in countering perception‐induced uncertainties remains unexplored. Yet, these methods are ideally suited to address disturbances from model mismatch and non‐Gaussian perception errors.

 In this paper, we present a computationally-efficient perception-driven tube-based MPC that 1) is formalized using linear programming, rather than quadratic programming, and 2) departs from the common assumption of Gaussian perception noise \cite{al2020accuracy, wei2023multi}, which has been shown to inadequately capture the complexities of real-world environments \cite{rahimi2022robust, maharmeh2025comprehensive}. To address this challenge, we propose a tube-based MPC framework that ensures constraint satisfaction while explicitly accounting for process and measurement noise with nonzero means.

We employ \emph{constrained zonotopes} \cite{scott2016constrained} as a unified representation for both uncertainty and invariant control sets. This enables efficient reachability analysis and allows the MPC optimization to be formulated as a linear program (LP), reducing computational overhead for onboard deployment by defining the cost function as a sum of Minkowski functions over zonotopic sets. To mitigate the dead-beat or idle behavior often observed in LP-based MPC schemes \cite{saffer2004analysis, rao2000linear}, we adopt the \(\infty/\infty\) cost formulation \cite{campo1989model}. Specifically, a competence-aware solution is determined at each step to balance accuracy with control effort. Furthermore, drawing on \cite{rakovic2012minkowski}, the terminal cost is systematically computed using a Minkowski–Lyapunov approach and is complemented by a terminal invariant zonotope to ensure closed-loop asymptotic stability.

\begin{figure*}[ht]
    \centering
    \includegraphics[height=0.2\textheight]{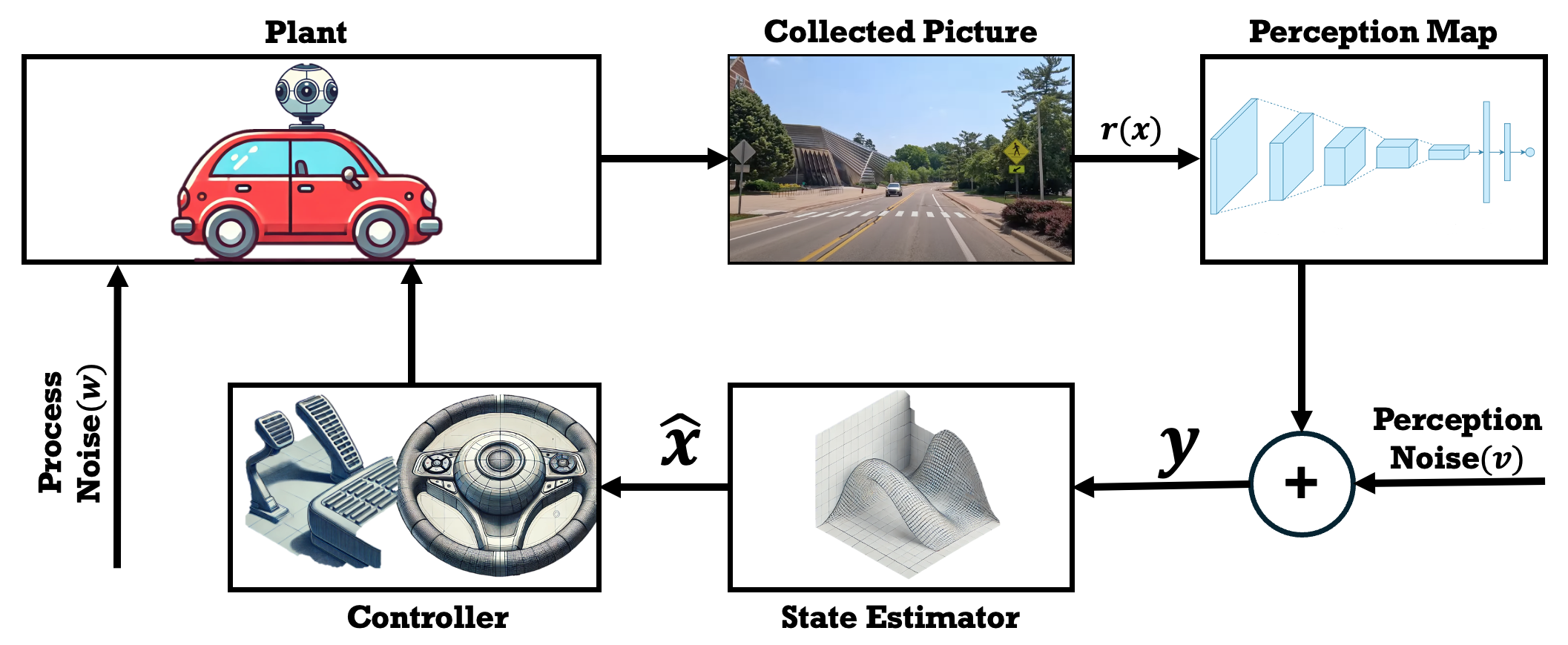}
        \caption{Block diagram of the closed-loop control architecture featuring a perception-based observer. A camera collects images of the plant, which are processed by the perception map to produce measurements subject to perception noise. These measurements, together with process noise, feed into the state estimator to generate the estimated state \(\hat{x}\). The controller then uses \(\hat{x}\) to compute the control inputs for the plant.}
     \label{Fig:pipline}
\end{figure*}

By integrating a CNN-based perception model for state estimation with a robust zonotopic observer and an LP-based MPC framework (Fig.~\ref{Fig:pipline}), and modeling perception noise as a zonotopic set, our approach addresses uncertainty and robustness challenges in perception-driven control while enabling faster onboard computation for autonomous vehicles \cite{rao2000linear}. A dataset is collected to train a novel, lightweight model—\texttt{RobotPerceptionNet} for regressing the physical state of a Husarion ROSbot XL (Husarion, Kraków, Poland). Both simulations and real-world experiments on an omnidirectional robot demonstrate that our method \textit{Perception-MPC} enhances safety and control performance compared to conventional approaches that rely on simplistic Gaussian noise assumptions.

\section{Preliminaries} 
This section presents the fundamentals of set-based operations, with a focus on constrained zonotopes—a computationally efficient representation of polytopes. Additionally, we review a methodology for ensuring the containment of one constrained zonotope within another, which is useful for designing an error-bounded observer in systems with zonotopic sensor noise.  
\subsection{Notations} In this paper, we use the symbols ${I}$, $\mathbb{N}$, $\mathbb{R}$, and $\mathbf{1}$ to denote the identity matrix, the set of natural numbers, the set of real numbers, and vector of ones, respectively. The ball \( \mathcal{B}_{\rho}(x_d) \) is defined as \( \mathcal B_{\rho}(x_d) := \{x \mid \Vert x - x_d \Vert \leq \rho\} \). Vectors are denoted in boldface, e.g., \(\mathbf{u} = [u_1, u_2, \dots, u_n]^\top\).

\subsection{Set Representations and Properties}
A polytope is characterized as the intersection of half-spaces.  
\begin{definition}\label{def:polytope}(Polytope)
   A polytope \( \mathcal{H} \subset \mathbb{R}^n \) is represented as the intersection of half-spaces and is formally defined as  
\begin{align}
    \mathcal{H} := \{x \in \mathbb{R}^n \mid Q x \leq q\}
\end{align}  
where \( Q \in \mathbb{R}^{m \times n} \) is a matrix and \( q \in \mathbb{R}^m \) is a vector, both used for characterizing the set of half-spaces. The parameter \( m \) denotes the total number of defining half-spaces. In this work, the notation \( \langle Q, q \rangle_H \) is adopted to represent \( \mathcal{H} \).  
\end{definition}
A zonotope represents a specific class of polytopes characterized by central symmetry. 
\begin{definition}\label{def:zonotope}(Zonotope)
    A zonotope is represented as  
\begin{align}
    \mathcal{Z} := \{x \in \mathbb{R}^n \mid x = c + G \bm{\alpha}, |\bm{\alpha}| \leq \mathbf{1} \}
\end{align}  
where \( c \in \mathbb{R}^n \) is the center, and \( G \in \mathbb{R}^{n \times \sigma(\mathcal{Z})} \) serves as the generator matrix, with \( \sigma(\mathcal{Z}) \) representing the total number of generators. The absolute value constraint is applied element-wise. The notation \( \langle c, G \rangle_{{Z}} \) is adopted to represent \( \mathcal{Z} \).  
\end{definition}

Zonotopes are a symmetric subclass of polytopes. Constrained zonotopes generalize this representation, removing the central symmetry requirement while preserving low computational complexity. 
\begin{definition}\label{def:Czonotope}(Constrained Zonotope)
    A constrained zonotope is defined by a center \( c \in \mathbb{R}^n \) and a generator matrix \( G \in \mathbb{R}^{n \times \sigma(\mathcal{Z}_c)} \) as  
\begin{align}\label{eq:Czonotope}
    \mathcal{Z}_c := \{x \in \mathbb{R}^n \mid x = c + G \bm{\alpha}, \; |\bm{\alpha}| \leq \mathbf{1}, \; F\bm{\alpha} = \theta\}
\end{align}  
where \( \sigma(\mathcal{Z}_c) \) denotes the total number of generators and \( F \in \mathbb{R}^{n_c \times \sigma(\mathcal{C})} \) and \( \theta \in \mathbb{R}^{n_c} \) where $n_c$ is the number of constraints. Similar to zonotopes, the notation \( \langle c, G, F, \theta \rangle_{{ Z_c}} \) is used to represent \( \mathcal{Z}_c \).  
\end{definition}

\begin{rem}\label{rem:H_rep}
Let 
\(
\mathcal{Z}_c = \langle c, G, F, \theta \rangle_{Z_c}
\)
be a constrained zonotope in~$\mathbb{R}^n$.  To obtain a polytope with \emph{H-representation} equivalent of $\mathcal{Z}_c$, i.e.\ find matrix $Q$ and vector~$q$ such that
\(
\mathcal{Z}_c \equiv\mathcal{H},
\)
one may proceed by projecting out the auxiliary variable~$\bm\alpha$ from the augmented polytope
\[
\begin{bmatrix}
- I_{\sigma(\mathcal{Z}_c)} & 0 \\[3pt]
  I_{\sigma(\mathcal{Z}_c)} & 0 \\[3pt]
  F   & 0 \\[3pt]
  G   & -I_n \\[3pt]
 -G   &  I_n
\end{bmatrix}
\begin{bmatrix}\bm\alpha\\ x\end{bmatrix}
\;\le\;
\begin{bmatrix}
1_{\sigma(\mathcal{Z}_c)} \\[3pt]
1_{\sigma(\mathcal{Z}_c)} \\[3pt]
\theta   \\[3pt]
-\,c \\[3pt]
c
\end{bmatrix}
\!.
\]
Eliminating $\bm\alpha$ via Fourier–Motzkin elimination \cite{dantzig1972fourier} (or any polytope‐projection algorithm) yields a finite set of linear inequalities in~$x$ alone, which is the equivalent polytope with \emph{H-representation} of a constrained zonotope $\mathcal{Z}_c$. \end{rem}

\begin{definition}\label{def:ellipsoid}(Ellipsoidal set)
  An ellipsoidal set \(\mathcal{E}\subset\mathbb{R}^n\) is defined as
\(
    \mathcal{E} := \{\,x\in\mathbb{R}^n \mid x^\top P^{-1} x \le 1\},
\)
  where \(P\in\mathbb{R}^{n\times n}\) is symmetric and positive definite. In this work, we adopt the notation \(\langle P\rangle_E\) to represent \(\mathcal{E}\).
\end{definition}

\begin{definition}\label{def:setBasedOperations}(Set-based Operations)
    The Minkowski sum of two sets, \( S_1 \) and \( S_2 \), is defined as  
\(
    S_1 \oplus S_2 := \{ s_1 + s_2 \mid s_1 \in S_1, s_2  \in S_2 \}.
\) 
Also, the linear transformation of a set \( S \) by a matrix \( L \in \mathbb{R}^{m \times n} \) is given by  
\(
    L \odot S := \{ Ls \mid s \in S \}.
\)
\end{definition}

 This class remains closed under Minkowski sum \( \oplus \) and linear transformation \( \odot \), as defined in Definition~\ref{def:setBasedOperations}. The results of these operations are computed as follows. Assume that \( \mathcal Z_c^1=\langle c_1, G_1, F_1, \theta_1 \rangle_{{ Z_c^1}} \) and \( \mathcal Z_c^2=\langle c_2, G_2, F_2, \theta_2 \rangle_{{ Z_c^2}} \). Then, 
\begin{subequations}
    \begin{align}
    \mathcal{Z}^1_{c} &\oplus \mathcal{Z}^2_{c} := \langle c_1 + c_2, \begin{bmatrix}
        G_1 & G_2
    \end{bmatrix}, \begin{bmatrix}
        F_1 & 0 \\ 0 & F_2
    \end{bmatrix}, \begin{bmatrix}
        \theta_1 \\ \theta_2
    \end{bmatrix} \rangle_{Z_c} \label{eq:MinkSum},\\
    L &\odot \mathcal{Z}^1_{c} := \langle Lc_1,LG_1, F_1, \theta_1 \rangle_{{Z}_c}. \label{eq:linTransform}
\end{align}
\end{subequations}

\subsection{Containing Constrained Zonotopes}
The following lemma, adapted from \cite{gheorghe2024computing}, formulates a linear programming problem that facilitates the efficient incorporation of constrained zonotope containment into a convex optimization framework.   
\begin{lem}\label{lem:containmentZonotope}\cite{gheorghe2024computing}
    Let \( \mathcal Z_c^1=\langle c_1, G_1, F_1, \theta_1 \rangle_{{ Z_c^1}} \) and \( \mathcal Z_c^2=\langle c_2, G_2, F_2, \theta_2 \rangle_{{ Z_c^2}} \) be constrianed zonotpes. \( \mathcal{Z}^1_{c} \) is contained within \( \mathcal{Z}^2_{c} \), denoted as \( \langle c_1, G_1, F_1, \theta_1 \rangle_{{Z}_c} \subseteq \langle c_2, G_2, F_2, \theta_2 \rangle_{{Z}_c} \), if there exist \( \Pi \in \mathbb{R}^{\sigma(\mathcal{Z}_c^2) \times \sigma(\mathcal{Z}_c^1)} \), \( H \in \mathbb{R}^{n^2_{c} \times n^1_{c}} \), and \( \gamma \in \mathbb{R}^{\sigma(\mathcal{Z}_c^2)} \) that satisfy the following conditions  
\begin{subequations}
    \begin{align}
        G_1 &= G_2 \Pi, \\
        H F_1 &= F_2 \Pi, \\
        H\theta_1 &= \theta_2 + F_2 \gamma, \\
        c_2 - c_1 &= G_2 \gamma, \\
        \left|  \Pi \right| \mathbf{1} + \left| \gamma  \right|  &\leq \mathbf{1}.
    \end{align}
\end{subequations}  
where \( n^1_{c} \) and \( n^2_{c} \) represent the number of constraints in \( \mathcal{Z}^1_{c} \) and \( \mathcal{Z}^2_{c} \), respectively. 
\end{lem}

\subsection{Minkowski Function}
\begin{definition}[Minkowski Function]\label{def:MinkowskiFunction}
Let \(\mathcal{S} \subset \mathbb{R}^n\) be a proper \(C\)-set. The \emph{Minkowski function} \(g(\mathcal{S},\cdot)\) is defined for all \(x\in\mathbb{R}^n\) by
\(
g(\mathcal{S},x) := \min\{\lambda \ge 0 \mid x \in \lambda \mathcal{S}\}.
\)
\end{definition}

\begin{definition}[Support Function]
Let \(\mathcal{S} \subset \mathbb{R}^n\) be a closed, non-empty, convex set. The \emph{support function} \(h(\mathcal{S},\cdot)\) is defined for all \(x\in\mathbb{R}^n\) by
\(
h(\mathcal{S},x) := \sup\{ x^\top y \mid y \in \mathcal{S} \}.
\)
\end{definition}
\begin{definition}[Polar Set] \label{def:polarSet}
For a set \(\mathcal{S} \subset \mathbb{R}^n\) with \(0 \in \mathcal{S}\), its \emph{polar set} is defined by
\(
\mathcal{S}^\star := \{ x \in \mathbb{R}^n \mid y^\top x \leq 1 \text{ for all } y \in \mathcal{S} \}.
\)
\end{definition}
\begin{rem}\label{rem:usefulProp}
 For any nonempty, closed, and convex sets \( \mathcal{S}, \mathcal{S}_1, \mathcal{S}_2 \subseteq \mathbb{R}^n \) and any matrix \( M \in \mathbb{R}^{p\times n} \), the following hold \cite{rakovic2020polarity}
\begin{itemize}  
    \item Set inclusion is equivalent to a support function inequality  
    \( \mathcal{S}_1 \subseteq \mathcal{S}_2 \iff \; h(\mathcal{S}_1, x) \leq h(\mathcal{S}_2, x), \quad \forall x \in \mathbb{R}^n\)
    \item The support function of the Minkowski sum satisfies  
    \( \; h(\mathcal{S}_1 \oplus \mathcal{S}_2, x) = h(\mathcal{S}_1, x) + h(\mathcal{S}_2, x),\quad \forall x \in \mathbb{R}^n  \)  
    \item The support function under linear transformation satisfies  
    \( \; h(M\mathcal{S}, y) = h(\mathcal{S}, M^\top y),\quad \forall y \in \mathbb{R}^p  \)  
    \item If \( \mathcal{S} \) is a proper \( C \)-set in \( \mathbb{R}^n \), its polar set \( \mathcal{S}^\star \) is also a proper \( C \)-set, and its bipolar set satisfies \( (\mathcal{S}^\star)^\star = \mathcal{S} \).  
    \item The Minkowski function \( g(\mathcal{S}, .) \) of a proper \( C \)-set \( \mathcal{S} \in  \mathbb{R}^n \) is equal to the support function  of its polar set \( \mathcal{S}^\star \) \( h(\mathcal{S}^\star, .) \).  
\end{itemize}  
\end{rem}

\section{Problem Setting and Formulation}
Consider a linear time-invariant discrete-time system as 
\begin{subequations}\label{eq:system}
\begin{align}
    x(k+1) &= A x(k) + B u(k) + w(k), \\
    o(k) &= r(x(k)), \label{eq:perceptionBased}
\end{align}  
\end{subequations}
where \( x \in \mathbb{R}^n \), \( u \in \mathbb{R}^m \), \( w \in \mathbb{R}^n \), and \( o \in \mathbb{R}^M \) represent the system state, control input, process noise, and observation, respectively. The system dynamics, characterized by \( (A, B) \), are assumed to be known. However, the observation process is governed by an unknown \textit{generative model} \( r \), which is nonlinear and potentially high-dimensional.  For example, consider a camera mounted on the dashboard of a car navigating a road. In this case, the observations at each time step \( o(k) \) are the captured images, while the function \( r \) maps the system’s state variables, such as position and velocity, to these visual observations. 
\begin{assumption}\label{ass:StateBound}
   The state of the system, governed by \eqref{eq:system}, is bounded by a convex set \( \mathcal{S}_{x} \), which is assumed to be represented as a constrained zonotope \( \mathcal{Z}^x_c = \langle c_x, G_x, F_x, \theta_x \rangle_{{Z}_c} \), where \( c_x \in \mathbb{R}^n \), \( G_x \in \mathbb{R}^{n \times \sigma(\mathcal{Z}^x_c)} \), \( F_x \in \mathbb{R}^{n^x_{c} \times \sigma(\mathcal{Z}^x_c)} \), and \( \theta_x \in \mathbb{R}^{n^x_{c}} \).  
\end{assumption}
\begin{assumption}\label{ass:InputBound}
   The input of the system, governed by \eqref{eq:system}, is bounded by a convex set \( \mathcal{S}_{u} \), which is assumed to be represented as a constrained zonotope \( \mathcal{Z}^u_c = \langle c_u, G_u, F_u, \theta_u \rangle_{{Z}_c} \), where \( c_u \in \mathbb{R}^m \), \( G_u \in \mathbb{R}^{m \times \sigma(\mathcal{Z}^x_u)} \), \( F_u \in \mathbb{R}^{n^u_{c} \times \sigma(\mathcal{Z}^u_c)} \), and \( \theta_u \in \mathbb{R}^{n^u_{c}} \).  
\end{assumption}
Assumptions~\ref{ass:StateBound} and \ref{ass:InputBound} stem from inherent system and environmental limitations, are non-restrictive, and align naturally with the system's safety constraints.
\begin{assumption}\label{ass:Contr}
    The pair $(A, B)$ is controllable. 
\end{assumption}
\begin{assumption}\label{ass:zonotopicProcessnoise}
    The process noise $w(k)$ is unknown but confined within a constrained zonotopic set, defined as \( \mathcal{Z}^w_c := \langle c_{w}, G_{w}, F_w, \theta_w \rangle_{{Z}_c} \subset \mathbb{R}^n \), ensuring that \( w(k) \in \mathcal{Z}^w_c\) for all \( k \geq 0 \) where $G_{w} \in \mathbb{R}^{n \times \sigma(\mathcal{Z}^w_c)}$, $c_{w} \in \mathbb{R}^n$, $F_w \in \mathbb{R}^{n_c^w \times \sigma(\mathcal{Z}^w_c)}$, and $\theta_w \in \mathbb{R}^{n_c^w}$.  
\end{assumption}

Before formulating the robust MPC problem, it is necessary to establish a structure that explicitly accounts for perception-induced noise and the corresponding observer design. By bounding the effects of process and measurement uncertainties within this framework, the problem formulation ensures that state and input trajectories remain within safety-critical constraints, enabling a tractable and robust control design in the subsequent subsections.

\subsection{Noise Characteristics, State Decomposition, and Bounded Sets}
\subsubsection{Perception Map}
Assume the existence of a \textit{perception map} \( \mathcal{E} \) that provides an imperfect prediction of partial state information, defined as \( \mathcal{E}(o(k)) \), where $o(k)$ is the output of the perception module at time $k$. Using this map, we introduce a new measurement model where \( \mathcal{E} \) acts as a noisy sensor
\begin{align}\label{eq:output}
    y(k) = \mathcal{E}(o(k)) = C x(k) + v(k),
\end{align}
with \( C \in \mathbb{R}^{\ell \times n} \) as a known matrix, \( v(k) \in \mathbb{R}^\ell \) representing prediction noise, and \( y(k) \in \mathbb{R}^\ell \) as the output. A perception model trained on a limited dataset may exhibit increased noise when the system operates in regions underrepresented in the training data \cite{smyl2021learning}. As illustrated in Fig.~\ref{Fig:pipline}, \( \mathcal{E} \) is implemented as a neural network trained via supervised learning on a dataset \( \mathcal{X} = \{(o_i, x_i) \}_{i=1}^{N} \), densely sampled around the system’s operating point \( x_d \), \cite{dean2021certainty}. Here, the perception noise \( v(k) \) captures the uncertainty in the mapping and is state-dependent, being smaller near the training data. We adopt the robust perception-based system model from \cite{dean2021certainty, dean2020robust} to establish robustness guarantees. Specifically, with effective training, the perception noise \( v(k) \) around \( x_d \) remains bounded, as formalized by a modified version of the assumption in.

\begin{assumption}\label{ass:noiseNearOperation}  
There exists a ball \( \mathcal{B}_\rho(x_d) \) of radius \( \rho \) centered at the operating point \( x_d \) such that the perception error satisfies 
\(
   \Vert \mathcal{E}(o(k)) - Cx(k) \Vert \leq \mu_v,\quad\quad  \forall x(k) \in \mathcal{B}_\rho(x_d)
\)
Equivalently, 
\(
   \Vert v(k) \Vert_{\infty} \leq \mu_v,\quad\quad  \forall x(k) \in \mathcal{B}_\rho(x_d),\,\,\forall \,k\geq 0
\)
\end{assumption}  

The sensor noise bound described in Assumption~\ref{ass:noiseNearOperation} is formally represented as a zonotopic set to capture its behavior. In \cite{al2020accuracy}, sensor noise is assumed to be Gaussian and independent and identically distributed (i.i.d.); however, this assumption is not realistic in practical applications. In practice, the sensor noise distribution depends on the system state and varies with its distance from the training data, making a Gaussian assumption unsuitable. To address this, we introduce the following assumption regarding the sensor noise.

\begin{assumption}\label{ass:zonotopicnoise}
    The sensor noise $v(k)$ is confined within a constrained zonotopic set, defined as \( \mathcal{Z}^v_c := \langle c_{v}, G_{v}, F_v, \theta_v \rangle_{{Z}_c} \subset \mathbb{R}^\ell \), ensuring that \( v(k) \in \mathcal{Z}^v_c\) for all \( k \geq 0 \) where $G_{v} \in \mathbb{R}^{\ell \times \sigma(\mathcal{Z}^v_c)}$, $c_{v} \in \mathbb{R}^\ell$, $F_v \in \mathbb{R}^{n_c^v \times \sigma(\mathcal{Z}^v_c)}$, and $\theta_v \in \mathbb{R}^{n_c^v}$.  
\end{assumption}
\subsubsection{State Estimator}
The dynamics of the state estimator, incorporating a static gain \( L \in \mathbb{R}^{n \times \ell} \), are formulated to estimate the system state in \eqref{eq:system} based on the available measurements \eqref{eq:output}:
\begin{align}\label{eq:estimator}
    \hat{x}(k+1) = A \hat{x}(k) + B u(k) + L \big[ y(k) - C \hat{x}(k) \big],
\end{align}
where \( \hat{x}(k) \) denotes the estimated state at time \( k \), and the estimation error is defined as \( e(k) := x(k) - \hat{x}(k) \).
Thus, one has the following for the error dynamics
\begin{align} \label{eq:errorDynamics}
    e(k+1) = (A - LC) e(k) - Lv(k) + w(k),\quad  \forall k \geq 0. 
\end{align}
\begin{assumption}\label{ass:Observ}
    The pair $(A, C)$ is observable. 
\end{assumption}
\subsubsection{Nominal System and State Deviation}
\label{sec:nominal_system}
By eliminating \( w \) and \( v \) from \eqref{eq:system}, the nominal system is defined as
\begin{align} \label{eq:nominalSystem} 
    \bar{x}(k) = A\bar{x}(k) + B \bar{u}(k),  
\end{align}  
where \( \bar{x} \in \mathbb{R}^n \) denotes the nominal state and \( \bar{u} \in \mathbb{R}^m \) is the associated control input.  

To mitigate the effects of process and sensor noise, the control input is formulated as a combination of the nominal control input and a feedback term  
\begin{align}\label{eq:controlInput}
    u(k) = \bar{u}(k) + K \Tilde{x}(k),
\end{align}
where \( \Tilde{x}(k) = \hat{x}(k) - \bar{x}(k) \) represents the estimation deviation, and \( K \in \mathbb{R}^{m\times n} \) is the feedback gain. With control input \eqref{eq:controlInput}, the closed-loop observer state satisfies
\begin{align}\label{eq:estimationSystem}
    \hat{x}(k+1) = A\hat{x}(k) + B\bar{u}(k) + BK \Tilde{x}(k) + LCe(k) + Lv(k).
\end{align}
Subtracting \eqref{eq:nominalSystem} from \eqref{eq:estimationSystem} results in the difference between the observer state and the nominal system state dynamics as
\begin{align}\label{eq:deviationSystem}
    \Tilde{x}(k+1) = (A+BK)\Tilde{x}(k) + L(Ce(k) + v(k)).
\end{align}
 
\subsubsection{Robust Invariant Set Principles}
With these developments, we express the state deviations as \( \Tilde{x}(k) = \hat{x}(k) - \bar{x}(k) \) and observer error as \( e(k) = x(k) - \hat{x}(k) \). Substituting these into the state equation yields  
\begin{align}\label{eq:stateComplete}
    x(k) = \bar{x}(k) + e(k) + \Tilde{x}(k).
\end{align}  
The key idea is that if both \( e(k) \) and \( \Tilde{x}(k) \) are properly bounded, in sets $\mathcal{S}_e$ and $\mathcal{S}_{\Tilde{x}}$, respectively, we can select an appropriate initial nominal state \( \bar{x}_0 \) and a sequence of nominal inputs \( \bar{\textbf{u}} \) to ensure that the actual system adheres to the state and input constraints defined in Assumptions~\ref{ass:StateBound} and \ref{ass:InputBound}. To ensure that the estimation error \( e(k) \) and state deviation \( \tilde{x}(k) \) remain within predefined bounds, we introduce the following key definitions.

\begin{definition}[\cite{blanchini2008set}]\label{def:robustInvSet}
    A set \( \mathcal{S} \) is a Robust Invariant Set (RIS) if \( x(k) \in \mathcal{S},\,\,\forall k \geq 0 \)  for any initial condition \( x_0 \in \mathcal{S} \) with a dynamic under \textit{uncertainties}.  
\end{definition}

To construct an RIS, we leverage the concept of \(\lambda\)-contractivity.
\begin{definition}[\cite{blanchini2008set}]\label{def:ContractiveSet}
    A set \( \mathcal{S} \) is \(\lambda\)-contractive if for a given \( \lambda \in (0,1] \) and \( x(k) \in \mathcal{S} \), the next state \( x(k+1)\in \lambda \mathcal {S}  \).   
\end{definition}

By ensuring that the bound sets $\mathcal{S}_e$ and $\mathcal{S}_{\tilde{x}}$, for the process noise \( \mathcal{Z}^w \) and sensor noise \( \mathcal{Z}^v \), are \(\lambda\)-contractive sets, we not only prevent \( e(k) \) and \( \Tilde{x}(k) \) from growing unbounded but also guarantee convergence to the actual state at a rate of at least \( \lambda \).  

The following lemma establishes the relationship between a constrained zonotopic set and its \( \lambda \)-contracted counterpart, as defined in Definition~\ref{def:ContractiveSet}.  
\begin{lem}\label{lem:lambdaContractedZonotope}
Let 
\(
\mathcal{Z}_c = \langle c, G, F, \theta\rangle_{Z_c}
\)
be a constrained zonotope in $\mathbb{R}^n$, with dimension following \eqref{eq:Czonotope}. For any scalar $\lambda\in (0,1]$, define the contracted set about the center $c$ as
\(
\mathcal{Z}_{\lambda c} = \langle c, \lambda G, F, \theta\rangle_{Z_c}\). 
Then, every point in $\mathcal{Z}_{\lambda c}$ lies in the line segment between the center $c$ and a corresponding point in $\mathcal{Z}_{c}$, and the 'size' of $\mathcal{Z}_{c}$ is uniformly scaled by $\lambda$ in all directions of the generator, while the constraint $F \bm{\alpha} = \theta$ remains unchanged.
\end{lem}
\textbf{Proof.}
Let $x \in \mathcal Z_c$. Then there exists some $\bm{\alpha}$ satisfying
\(
|\bm{\alpha}| \leq \textbf{1} \quad \text{and} \quad F\bm{\alpha} = 	\theta
\)
such that
\(
x = c+G\bm{\alpha}.
\)
Consider the corresponding point in the contracted set
\(
x_\lambda = c+\lambda G\bm{\alpha}.
\)
Since $\lambda \in (0,1]$, we can write
\(
x_\lambda = c + \lambda \left( G\bm{\alpha} \right) = c + \lambda (x - c).
\)
Thus, $x_\lambda$ is obtained by scaling the displacement $(x-c)$ by $\lambda$ while keeping the center $c$ fixed. In other words, each point in $\mathcal{Z}_{\lambda c}$ is the contraction of a point in $\mathcal{Z}_{c}$ toward $c$ by the factor $\lambda$. 
Moreover, the constraint on $\bm{\alpha}$ remains unchanged since the same $\bm{\alpha}$ that satisfies $|\bm{\alpha}| \leq \textbf{1}$ and $F\bm{\alpha} = \theta$ is used in both definitions of $\mathcal{Z}_{c}$ and $\mathcal{Z}_{\lambda c}$. Thus, the slicing of the unit hypercube $\{\bm{\alpha} : |\bm{\alpha}| \le \textbf{1}\}$ by the equality constraint is invariant under the contraction. 
$\hfill \Box$

The following remark reviews and formally states the requirements to formalize Problem~\ref{problem.1} using the state error and deviation sets.
\begin{rem}\label{rem:setConditons}
The initial observer error set \( \mathcal{S}_e \) and state deviation set \( \mathcal{S}_{\tilde{x}} \) serve as design parameters that influence the overall performance of the system. Based on established developments, the following conditions hold:
1. Since the estimation error is defined as \( e(k) = x(k) - \hat{x}(k) \), ensuring that the observer state estimate satisfies \( \hat{x} \in \mathcal{S}_x \ominus \mathcal{S}_e \) guarantees that the actual state remains within the constraint set \( \mathcal{S}_x \), i.e., \( x \in \mathcal{S}_x \).

2. Considering the control input structure in \eqref{eq:controlInput}, the nominal control input \( \bar{u} \) must satisfy the tightened constraint \( \bar{u} \in \mathcal{S}_u \ominus K \mathcal{S}_{\tilde{x}} \) to ensure feasibility.

3. Based on \eqref{eq:stateComplete}, to maintain the constraint satisfaction for the true state, the nominal state must satisfy \( \bar{x} \in \mathcal{S}_x \ominus (\mathcal{S}_e \oplus \mathcal{S}_{\tilde{x}}) \).

These conditions ensure that both state and control constraints remain satisfied under the influence of estimation and state deviation uncertainties.
\end{rem}

\begin{assumption}
    \label{ass:ErrorBound}
    The estimation error set \( \mathcal{S}_{e} \) is assumed to be represented as a constrained zonotope \( \mathcal{Z}^e_c = \langle c_e, G_e, F_e, \theta_e \rangle_{{Z}_c} \), where \( c_e \in \mathbb{R}^n \), \( G_e \in \mathbb{R}^{n \times \sigma(\mathcal{Z}^e_c)} \), \( F_e \in \mathbb{R}^{n^e_{c} \times \sigma(\mathcal{Z}^e_c)} \), and \( \theta_e \in \mathbb{R}^{n^e_{c}} \).
\end{assumption}

\begin{assumption}
    \label{ass:DeviationBound}
    The state deviation set \( \mathcal{S}_{\tilde{x}} \) is assumed to be represented as a constrained zonotope \( \mathcal{Z}^{\tilde{x}}_c = \langle c_{\tilde{x}}, G_{\tilde{x}}, F_{\tilde{x}}, \theta_{\tilde{x}} \rangle_{{Z}_c} \), where \( c_{\tilde{x}} \in \mathbb{R}^n \), \( G_{\tilde{x}} \in \mathbb{R}^{n \times \sigma(\mathcal{Z}^{\tilde{x}}_c)} \), \( F_{\tilde{x}} \in \mathbb{R}^{n^{\tilde{x}}_{c} \times \sigma(\mathcal{Z}^{\tilde{x}}_c)} \), and \( \theta_{\tilde{x}} \in \mathbb{R}^{n^{\tilde{x}}_{c}} \).
\end{assumption}

\subsection{Problem Formulation} 
\label{sec:problemformulation}
 In the following, we formulate a tube-based robust MPC problem that is tailored to a perception-driven observer and accelerated by expressing both the stage cost and terminal costs as Minkowski functions, thereby casting the entire optimization as an LP. 

\begin{problem}\label{problem.1}
Consider system~\eqref{eq:system} under Assumptions~\ref{ass:StateBound}–\ref{ass:DeviationBound}. At each time step \(k \ge 0\) with an initial state \(x_0\), solve the finite-horizon optimization to obtain the nominal input sequence
\(
\bigl\{\bar u(k\!\mid\!k),\,\bar u(k+1\!\mid\!k),\,\dots,\,\bar u(k+N-1\!\mid\!k)\bigr\}.
\)
Only the first element \(\bar u(k\!\mid\!k)\) is then applied to the plant via \eqref{eq:controlInput}
\begin{subequations}\label{eq:MPCformulation}
    \begin{align}
    &J^{*} := \min_{\bar{\textbf{x}}, \bar{\textbf{u}}} \Sigma^{k+N-1}_{j = k} \ell(j|k) + \ell_f(k+N|k), \label{eq:cost}\\
    \text{s.t.} \quad & \forall j=[k,k+N-1], \notag \\
    &\bar{x}(j+1|k) = A\bar{x}(j|k) + B\bar{u}(j|k) \label{eq:nominalDynamics}\\
    &\bar{x}(j|k) \in \mathcal{S}_{\bar{x}} := \mathcal{S}_x \ominus (\mathcal{S}_e \oplus \mathcal{S}_{\tilde{x}}) \label{eq:nominalXconst}\\
    &\bar{u}(j|k) \in \mathcal{S}_{\bar{u}} := \mathcal{S}_u \ominus K \mathcal{S}_{\tilde{x}} \label{eq:nominalUconst}\\
    &{e}(k) = [{x}(k) - \hat{x}(k)] \in \mathcal{S}_{e} \label{eq:observerCondition}\\
    &\tilde{x}(k) = [\hat{x}(k) - \bar{x}(k|k)] \in \mathcal{S}_{\tilde{x}} \label{eq:deviationCondition}\\
    &\bar{x}(k+N|k) \in \mathcal{S}_{f},\label{eq:finalSetCons}\\
    &\bar{x}(k | k) = x_0.
    \end{align}
    where \(N\) denotes the prediction horizon. The stage cost is defined as \( \ell(j|k) = g(\mathcal{Q}, \bar{x}(j)) + g(\mathcal{R}, \bar{u}(j))\) and the terminal cost is given by \( \ell_f(k+N|k) = g(\mathcal{P}, \bar{x}(k+N))\), where \(g(\mathcal{S}, \cdot)\) is the Minkowski function as defined in Definition~\ref{def:MinkowskiFunction}. The sets \(\mathcal{Q}\), \(\mathcal{R}\), and \(\mathcal{P}\) are proper C-sets in \(\mathbb{R}^n\), \(\mathbb{R}^m\), and \(\mathbb{R}^n\), respectively. In this optimization problem, \(\mathcal{S}_f\) is the terminal constraint set, \(\mathcal{S}_e\) and \(\mathcal{S}_{\tilde{x}}\) denote the estimation error and state deviation sets, and \(\mathcal{S}_x\) and \(\mathcal{S}_u\) represent the state and input constraints, respectively.
   
\end{subequations}

\end{problem}

To address Problem~1, the constraints \eqref{eq:nominalDynamics}–\eqref{eq:nominalUconst} are derived from the known system model \((A,B)\). The safety sets \(\mathcal{S}_x\) and \(\mathcal{S}_u\) are tightened to \(\mathcal{S}_{\bar{x}} = \mathcal{S}_x \ominus \big(\mathcal{S}_e \oplus \mathcal{S}_{\tilde{x}}\big)\) and \(\mathcal{S}_{\bar{u}} = \mathcal{S}_u \ominus K\mathcal{S}_{\tilde{x}}\), once the design sets \(\mathcal{S}_e\) and \(\mathcal{S}_{\tilde{x}}\) are specified (Assumptions~\ref{ass:ErrorBound}–\ref{ass:DeviationBound}; Remark~\ref{rem:setConditons}) and the observer and feedback gains \(L\) and \(K\) are chosen such that \(\mathcal{S}_e\) and \(\mathcal{S}_{\tilde{x}}\) are robustly invariant or contractive (Theorems~\ref{thm:zonotopeObserver}–\ref{thm:zonotopeFeedbackGain}). Under these conditions, constraints \eqref{eq:observerCondition}–\eqref{eq:deviationCondition} are satisfied. Finally, Theorems~\ref{thm:finalSetComp}–\ref{thm:maxSize} establish a stabilizing terminal gain \(K_f\) and a maximal terminal set \(\mathcal{S}_f\) that fulfill both the MPC terminal conditions and the terminal constraint \eqref{eq:finalSetCons}.

\section{Robustly Bounding the Estimation Error and State Deviation}
 
In this section, we establish the following results to ensure that the constraints \eqref{eq:observerCondition} and \eqref{eq:deviationCondition} are satisfied by bounding the estimation error \( e(k) \) and state deviation \( \tilde{x}(k) \).  

\subsection{Bounding the Estimation Error}
\label{sub:StateError}

Although a Kalman gain is optimal for systems with white, zero‑mean Gaussian noise~\cite{grewal2014kalman}, real‑world measurements are often Coloured and biased, making those assumptions unrealistic.  Consequently, we relax the noise model in Assumptions~\ref{ass:zonotopicProcessnoise}–\ref{ass:zonotopicnoise}: the process noise \(w(k)\) and sensor noise \(v(k)\) need not be Gaussian, white, or zero mean.  Instead of embedding a Kalman gain in a Luenberger observer—an approach that can introduce bias and even instability under non‑ideal noise—we design the observer gain \(L\) to guarantee a bounded estimation error with a prescribed convergence rate.  The error bound is represented by a constrained zonotope, ensuring robustness against measurement disturbances with arbitrary mean and color.
 
\begin{lem}\label{lem:ZonotopicPropagation}
Consider the system dynamics \eqref{eq:system} and the estimator dynamics \eqref{eq:estimator}. Assume that the system state operates in the region $x(k) \in \mathcal{B}_\rho(x_d)$ and  Assumptions~\ref{ass:zonotopicProcessnoise}--\ref{ass:ErrorBound} hold. Suppose the estimation error at time step $k$, denoted by $e(k)$, is contained in the constrained zonotope 
\(
\mathcal{Z}^e_c(k) = \langle c^k_e, G^k_e, F^k_e, \theta^k_e \rangle_{{Z}_c},
\)
and the sensor noise $v(k)$ and process noise $w(k)$ are contained in the zonotopes $\mathcal{Z}_c^v$ and $\mathcal{Z}_c^w$ as defined in Assumptions~\ref{ass:zonotopicnoise} and \ref{ass:zonotopicProcessnoise}. Then, the error dynamics \eqref{eq:errorDynamics} imply that the estimation error at time step $k+1$ is contained in the constrained zonotope
\[
\mathcal{Z}^e_c(k+1) = \langle c^{k+1}_e, G^{k+1}_e, F^{k+1}_e, \theta^{k+1}_e \rangle_{{Z}_c},
\]
where the parameters evolve according to
\begin{subequations}\label{eq:errorPropagation}
\begin{align}
    c^{k+1}_e &= (A - LC)c^k_e - L c_{v} +c_w, \label{eq:centerPropagation}\\[1mm]
    G^{k+1}_e &= \begin{bmatrix} (A - LC)G^k_e & -L G_v & G_w \end{bmatrix}, \label{eq:genMatrixPropagation}\\[1mm]
    F^{k+1}_e &= \begin{bmatrix} F^k_e & 0 & 0 \\ 0 & F_v &0\\ 0 & 0 &F_w\end{bmatrix}, \label{eq:constraintMatrixPropagation}\\[1mm]
    \theta^{k+1}_e &= \begin{bmatrix} \theta^k_e \\ \theta_v \\\theta_w \end{bmatrix}. \label{eq:constraintBoundsPropagation}
\end{align}
\end{subequations}
\end{lem}

\textbf{Proof.} The proof follows directly from the estimation error dynamics in \eqref{eq:errorDynamics}. By applying the properties of linear transformations and Minkowski sums for zonotopes, as defined in \eqref{eq:MinkSum} and \eqref{eq:linTransform}, the propagation of the center, generator matrix, and the constraints is obtained immediately. 
 $\hfill \Box$

Building on Lemmas~\ref{lem:lambdaContractedZonotope} and \ref{lem:ZonotopicPropagation}, we now establish a bounded esimation error set that ensures the estimation error remains within a predefined constrained zonotopic boundary. Specifically, given an initial estimation error bound, we guarantee that the estimation error remains contained within this set at all times.   
\begin{thm}\label{thm:zonotopeObserver}
    Consider the system \eqref{eq:system} and the estimator \eqref{eq:estimator} under Assumptions \ref{ass:Contr}--\ref{ass:ErrorBound}. Suppose an initial zonotopic estimation error bound is available, represented by 
    \[
    \mathcal{Z}^e_c = \langle c_e, G_e, F_e, \theta_e \rangle_{Z_c}.
    \]
    Then, the set $\mathcal{Z}^e_c$ is contractive with a contraction ratio of $\lambda_L \in (0,1]$, if there exist matrices \(L \in \mathbb{R}^{n \times \ell}\) as the observer gain in the error dynamics \eqref{eq:errorDynamics}, \(\Pi \in \mathbb{R}^{\sigma(\mathcal{Z}^e_c) \times (\sigma(\mathcal{Z}^e_c) + \sigma(\mathcal{Z}_c^v) + \sigma(\mathcal{Z}_c^w))}\), \(H \in \mathbb{R}^{(n_c^e + n_c^v + n_c^w) \times n_c^e}\), and a vector \(\gamma \in \mathbb{R}^{\sigma(\mathcal{Z}_c^e)}\) satisfying
\begin{subequations}\label{eq:zonotopeObserver}
    \begin{align}
        \begin{bmatrix} (A - LC)G_e & -L G_v & G_w \end{bmatrix} &= \lambda_L G_e \Pi, \\
        H \begin{bmatrix} F_e & 0 & 0 \\ 0 & F_v &0\\ 0 & 0 &F_w\end{bmatrix} &= F_e \Pi, \\
        H\begin{bmatrix} \theta_e \\ \theta_v \\ \theta_w \end{bmatrix} &= \theta_e + F_e \gamma, \\
        c_e - \Big((A - LC)c_e - L c_{v} +c_w\Big) &= \lambda_L G_e \gamma, \\
        \left|  \Pi \right| \mathbf{1}+ \left| \gamma  \right|  &\leq \mathbf{1}.
    \end{align}
\end{subequations}    
\end{thm}
\textbf{Proof.}  
First, by applying Lemma~\ref{lem:ZonotopicPropagation}, one can compute the propagation of the initial estimation error set \(\mathcal{Z}^e_c\) under the error dynamics, resulting in a propagated error zonotope \(\mathcal{Z}_c^1\). Next, Lemma~\ref{lem:lambdaContractedZonotope} provides the \(\lambda\)-contracted version of the initial error set, denoted by \(\mathcal{Z}_c^2\). According to Definitions~\ref{def:robustInvSet} and \ref{def:ContractiveSet}, if the propagated error set \(\mathcal{Z}_c^1\) is contained within the contracted set \(\mathcal{Z}_c^2\), that is,
\(
\mathcal{Z}_c^1 \subseteq \mathcal{Z}_c^2,
\)
then the observer design problem to make $\mathcal{Z}_c^e$, RIS, is solved. This containment is done using Lemma~\ref{lem:containmentZonotope}, which leads to \eqref{eq:zonotopeObserver}. With this, the proof is complete. \(\hfill \Box\)

\subsection{Bounding the State Deviation}
Similar to Subsection~\ref{sub:StateError}, in this subsection, we bound the state deviation $\tilde{x}(t)$. 

\begin{lem}\label{lem:ZonotopicPropagationDeviation}
Consider the system dynamics \eqref{eq:system} and the state deviation dynamics \eqref{eq:deviationPropagation}. Assume that the system state operates in the region $x(k) \in \mathcal{B}_\rho(x_d)$ and Assumptions~\ref{ass:Contr}--\ref{ass:Observ} and \ref{ass:DeviationBound} hold. Suppose that the state deviation $\tilde{x}(k)$ at time step $k$ is contained in the constrained zonotope 
\(
\mathcal{Z}^{\tilde{x}}_c(k) = \langle c^k_{\tilde{x}}, G^k_{\tilde{x}}, F^k_{\tilde{x}}, \theta^k_{\tilde{x}} \rangle_{{Z}_c},
\)
and the sensor noise $v(k)$ and process noise $w(k)$ are contained in the zonotopes $\mathcal{Z}_c^v$ and $\mathcal{Z}_c^w$ as defined in Assumptions~\ref{ass:zonotopicnoise} and \ref{ass:zonotopicProcessnoise}. Let an observer gain $L$ be designed with Theorem~\ref{thm:zonotopeObserver} such that it makes the constrained zonotopic set $\mathcal{Z}_c^e$, RIS. Then, the state deviation dynamics \eqref{eq:deviationSystem} imply that the state deviation at time step $k+1$ is contained in the constrained zonotope
\(
\mathcal{Z}^{\tilde{x}}_c(k+1) = \langle c^{k+1}_{\tilde{x}}, G^{k+1}_{\tilde{x}}, F^{k+1}_{\tilde{x}}, \theta^{k+1}_{\tilde{x}} \rangle_{{Z}_c},
\)
where the parameters evolve according to
\begin{subequations}\label{eq:deviationPropagation}
\begin{align}
    c^{k+1}_{\tilde{x}} &= (A + BK)c^k_{\tilde{x}} + LC c_{e} + Lc_v, \label{eq:centerPropagationDev}\\[1mm]
    G^{k+1}_{\tilde{x}} &= \begin{bmatrix} (A + BK)G^k_{\tilde{x}} & L C G_e & LG_v \end{bmatrix}, \label{eq:genMatrixPropagationDev}\\[1mm]
    F^{k+1}_{\tilde{x}} &= \begin{bmatrix} F^k_{\tilde{x}} & 0 & 0 \\ 0 & F_e &0\\ 0 & 0 &F_v\end{bmatrix}, \label{eq:constraintMatrixPropagationDev}\\[1mm]
    \theta^{k+1}_{\tilde{x}} &= \begin{bmatrix} \theta^k_{\tilde{x}} \\ \theta_e \\\theta_v \end{bmatrix}. \label{eq:constraintBoundsPropagationDev}
\end{align}
\end{subequations}
\end{lem}

\textbf{Proof.} The proof is similar to the one of Lemma~\ref{lem:ZonotopicPropagation}.
 $\hfill \Box$

Building on Lemmas~\ref{lem:lambdaContractedZonotope} and \ref{lem:ZonotopicPropagationDeviation}, we derive a bounded state-deviation set that guarantees the deviation remains within a predefined constrained zonotopic boundary. Given an initial deviation bound, the state deviation is ensured to stay within this set for all time steps.

\begin{thm}\label{thm:zonotopeFeedbackGain}
    Consider the system \eqref{eq:system} and the state deviation \eqref{eq:deviationSystem} under Assumptions \ref{ass:Contr}--\ref{ass:Observ} and \ref{ass:DeviationBound}. Suppose an initial zonotopic state deviation bound is available, represented by 
    \(
    \mathcal{Z}^{\tilde{x}}_c = \langle c_{\tilde{x}}, G_{\tilde{x}}, F_{\tilde{x}}, \theta_{\tilde{x}} \rangle_{Z_c}.
    \)
    Let an observer gain $L$ be designed with Theorem~\ref{thm:zonotopeObserver} such that it makes the constrained zonotopic set $\mathcal{Z}_c^e$, RIS. Then, the set $\mathcal{Z}^{\tilde{x}}_c$ is contractive with a contraction ratio of $\lambda_{\tilde{x}} \in (0,1]$, if there exist matrices \(K \in \mathbb{R}^{m \times n}\) as a feedback gain in \eqref{eq:deviationSystem}, \(\Pi \in \mathbb{R}^{\sigma(\mathcal{Z}^{\tilde{x}}_c) \times (\sigma(\mathcal{Z}^{\tilde{x}}_c) + \sigma(\mathcal{Z}_c^e) + \sigma(\mathcal{Z}_c^v))}\), \(H \in \mathbb{R}^{(n_c^{\tilde{x}} + n_c^e + n_c^v) \times n_c^{\tilde{x}}}\), and a vector \(\gamma \in \mathbb{R}^{\sigma(\mathcal{Z}_c^{\tilde{x}})}\) satisfying
\begin{subequations}\label{eq:zonotopeDeviation}
    \begin{align}
        \begin{bmatrix} (A + BK)G_{\tilde{x}} & LC G_e & LG_v \end{bmatrix} &= \lambda_{\tilde{x}} G_{\tilde{x}} \Pi, \\
        H \begin{bmatrix} F_{\tilde{x}} & 0 & 0 \\ 0 & F_e &0\\ 0 & 0 &F_v\end{bmatrix} &= F_{\tilde{x}} \Pi, \\
        H\begin{bmatrix} \theta_{\tilde{x}} \\ \theta_e \\ \theta_v \end{bmatrix} &= \theta_{\tilde{x}} + F_{\tilde{x}} \gamma, \\
        c_{\tilde{x}} - \Big((A + BK)c_{\tilde{x}} + LC c_{e} + Lc_v\Big) &= \lambda_{\tilde{x}} G_{\tilde{x}} \gamma, \\
        \left|  \Pi \right| \mathbf{1}+ \left| \gamma  \right|  &\leq \mathbf{1}.
    \end{align}
\end{subequations}    
\end{thm}

\textbf{Proof.}  
The proof is similar to that of Theorem~\ref{thm:zonotopeObserver}. \(\hfill \Box\)

\section{Robust MPC Design for Systems with Perception-based Observers}
Thus far, constraints \eqref{eq:observerCondition} and \eqref{eq:deviationCondition} have been satisfied via our set containment theorems. For the remaining constraints in Problem~\ref{problem.1}, the nominal state \(\bar{\textbf{x}}\) and nominal control sequence \(\bar{\textbf{u}}\) are treated as decision variables. Given an initial \(\bar{x}(k|k)\) that meets \eqref{eq:nominalXconst}, the nominal control sequence must satisfy \eqref{eq:nominalUconst} while minimizing the cost \eqref{eq:cost} online. Hence, solving the optimization problem \eqref{eq:MPCformulation} with the remaining constraints, one has 
    \begin{align}
        (\bar{\textbf{x}}^*,\bar{\textbf{u}}^*) &= \arg\min_{\bar{\textbf{x}},\bar{\textbf{u}}} J^{*}. \label{eq:ansXU}
    \end{align}
Then, using \eqref{eq:controlInput} and the solution \eqref{eq:ansXU}, the model predictive control law at time \(k\) is given by
\begin{align}\label{eq:MPC_control_Input}
    u(k) = \bar{u}^*(k|k) + K\big(\hat{x}(k)-\bar{x}^*(k|k)\big),
\end{align}
where \(\bar{u}^*(k|k)\) is the first element of \(\bar{\textbf{u}}^*\).

\textbf{Terminal Cost Function Design:}
As for constraint \eqref{eq:finalSetCons}, to satisfy standard stabilizing conditions in MPC, the terminal cost \(\ell_f(\cdot)\) and terminal constraint set \(\mathcal{S}_f\) must fulfill
\begin{subequations}
\begin{align}
\ell_f((A+BK_f)x) + \ell(x, K_fx) \leq \ell_f(x) \; \; \forall x \in \mathcal{S}_f, \label{eq:secondCondition}
            \\(A+BK_f) \mathcal{S}_f \subseteq \mathcal{S}_f, \mathcal{S}_f \subseteq \mathcal{S}_{\bar{x}}, \text{and} \; K_f \mathcal{S}_f \subset \mathcal{S}_{\bar{u}}.\label{eq:firstCondition}
    \end{align}
\end{subequations}

Note that \( K_f \) is not necessarily equivalent to \( K \). To satisfy the condition \( \eqref{eq:secondCondition} \), it is crucial to develop a methodology to determine a feedback gain \( K_f \) that meets this condition, using the specified stage ($\ell(x, K_f x)$) and terminal ($\ell_f(x)$) costs. The following describes the approach to achieve this.

\begin{thm}\label{thm:finalSetComp}
   Let the terminal cost be defined as the Minkowski function \( \ell_f(x) = g(\mathcal{P}, x) \) associated with a proper \( C \)-set \(\mathcal{P} \subset \mathbb{R}^n\), while the stage cost is given by \( \ell(x, K_f x) = g(\mathcal{Q}, x) + g(\mathcal{R}, K_f x) \) with proper \( C \)-sets \(\mathcal{Q} \subset \mathbb{R}^n\) and \(\mathcal{R} \subset \mathbb{R}^m\). The feedback gain \(K_f \in \mathbb{R}^{m \times n}\), found from to \eqref{eq:finalSetCost}, satisfies \eqref{eq:secondCondition} if the polar sets \( \mathcal{P}^\star\), \( \mathcal{Q}^\star\), and \( \mathcal{R}^\star\) (represented by the constrained zonotopic sets \(\mathcal{Z}^{p^*}_c\), \(\mathcal{Z}^{q^*}_c\), and \(\mathcal{Z}^{r^*}_c\), respectively), with \(\Pi \in \mathbb{R}^{\sigma(\mathcal{Z}^{p^*}_c) \times (\sigma(\mathcal{Z}^{p^*}_c)+\sigma(\mathcal{Z}^{q^*}_c)+\sigma(\mathcal{Z}^{r^*}_c))} \), \(H \in \mathbb{R}^{(n_c^{p^*} + n_c^{q^*} + n_c^{r^*}) \times n_c^{p^*}}\), \(\gamma \in \mathbb{R}^{\sigma(\mathcal{Z}_c^{p^*})}\), satisfy the following constraints
    \begin{subequations} \label{eq:finalSetCost}
        \begin{align}
        \begin{bmatrix} (A + BK_f)^\top G_{p^*} & G_{q^*} & K_f^\top G_{r^*} \end{bmatrix} &=  G_{p^*} \Pi, \\
        H \begin{bmatrix} F_{p^*} & 0 & 0 \\ 0 & F_{q^*} &0\\ 0 & 0 &F_{r^*}\end{bmatrix} &= F_{p^*} \Pi, \\
        H\begin{bmatrix} \theta_{p^*} \\ \theta_{q^*} \\ \theta_{r^*} \end{bmatrix} = \theta_{p^*} + &F_{p^*} \gamma, \\
        c_{p^*} - \Big((A + BK_f)^\top c_{p^*} +  c_{q^*} + K_f^\top c_{r^*}\Big) &=  G_{p^*} \gamma, \\
        \left|  \Pi \right| \mathbf{1}+ \left| \gamma  \right|  &\leq \mathbf{1}
        \end{align}
    \end{subequations}
\end{thm}

\textbf{Proof.} By expressing condition \eqref{eq:secondCondition} and applying the properties of the Minkowski function and the support function of its polar set (see Remark~\ref{rem:usefulProp}), we obtain
\begin{align}
    &g(\mathcal{P},(A+BK_f)x) - g(\mathcal{P},x) \leq -g(\mathcal{Q},x) -g(\mathcal{R},K_fx), \notag \\
    &h(\mathcal{P}^\star,(A+BK_f)x) - h(\mathcal{P}^\star,x) \leq \notag \\ &\quad\quad\quad\quad\quad -h(\mathcal{Q}^\star,x) -h(\mathcal{R}^\star,K_fx), \notag \\
    &h((A+BK_f)^\top\mathcal{P}^\star,x) - h(\mathcal{P}^\star,x) \leq \notag \\ &\quad\quad\quad\quad\quad -h(\mathcal{Q}^\star,x) -h(K_f^\top \mathcal{R}^\star,x), \notag\\
    &h((A+BK_f)^\top\mathcal{P}^\star +\mathcal{Q}^\star+K_f^\top \mathcal{R}^\star,x)  \leq h(\mathcal{P}^\star,x), \notag \\
    &(A+BK_f)^\top\mathcal{P}^\star \oplus \mathcal{Q}^\star\oplus K_f^\top \mathcal{R}^\star \subseteq \mathcal{P}^\star. \label{eq:setIncFinal} 
\end{align}
To satisfy \eqref{eq:setIncFinal}, using Lemma~\ref{lem:containmentZonotope}, one has \eqref{eq:finalSetCost}. With that, the proof is concluded. 
\(\hfill \Box\)

\begin{rem}[Aspiration level via the terminal set $\mathcal{P}$ and optimal gain $K_f$]
Let $\mathcal{P}\subset\mathbb{R}^{n}$ be the proper $\mathcal{C}$‑set whose Minkowski gauge $\ell_{f}(x)=g(\mathcal{P},x)$ serves as the terminal cost in Theorem~\ref{thm:finalSetComp}. 
Because $g(.,.)$ is monotone with respect to set inclusion,  
\(
\mathcal{P}_{1}\subset\mathcal{P}_{2}\;\Longrightarrow\;
g(\mathcal{P}_{1},x)\ge g(\mathcal{P}_{2},x)\quad\forall x\in\mathbb{R}^{n}.
\)
Hence, shrinking $\mathcal{P}$ tightens the inequality~\eqref{eq:secondCondition}, whereas enlarging $\mathcal{P}$ relaxes the inequality~\eqref{eq:secondCondition}. This maximal set is termed the \emph{least conservative aspiration set}. It can be obtained by iteratively increasing the size of an initial candidate $\hat{\mathcal{P}}$ that satisfies \eqref{eq:secondCondition} via Theorem~\ref{thm:finalSetComp}. Additionally, one can find the \textit{optimal} terminal set gain $K_f$ using the iterative pipeline in \cite{rakovic2021minkowski} or an approximation of it based on linear quadratic regulator gain using the similarity between the linear and quadratic costs in \cite{saffer2004analysis}.
    \end{rem}

\begin{thm}\label{thm:maxSize}
Let the nominal closed‐loop matrix \(A_f := A + B K_f\) be given, where \(K_f\) satisfies the stabilizing condition of Theorem~\ref{thm:finalSetComp}.  Suppose the tightened state \(\mathcal{S}_{\bar x}\) and input constraint sets \(\mathcal{S}_{\bar u}\) are polytopes in \(\mathbb R^n\) and \(\mathbb R^m\), respectively, shown by \(\langle Q_{\bar x}, q_{\bar x}\rangle_{H}\) and \(\langle Q_{\bar u}, q_{\bar u}\rangle_{H}\) obtained from original constrained zonotopic sets \(\mathcal{Z}^{\bar x}_c\) and \(\mathcal{Z}^{\bar u}_c\) via Remark~\ref{rem:H_rep}.  Define the terminal set using an ellipsoidal candidate
\(
S_f = \langle P \rangle_E , P\succ0.
\)
Then, for a contraction ratio $\lambda_{{f}} \in (0,1]$ the unique maximizer \(P^\star\) of the convex program
\begin{subequations}
\begin{align}
\max_{P\succ0}\;&\log\det P,\\
\text{s.t.}\quad 
&A_f\,P\,A_f^\top \preceq \lambda_{{f}}\,P,\\
&Q_x\,P\,Q_x^\top \preceq \mathrm{diag}(q_x),\\
&Q_u\,K_f\,P\,K_f^\top\,Q_u^\top \preceq \mathrm{diag}(q_u),
\end{align}
\end{subequations}
defines the maximum‐volume ellipsoid satisfying
\begin{enumerate}
  \item \((A+BK_f)\,S_f \subseteq S_f\) with contraction factor \(\lambda_f\),
  \item \(S_f \subseteq \mathcal{S}_{\bar x}\) and \(K_f\,S_f \subseteq \mathcal{S}_{\bar u}\).
\end{enumerate}
\end{thm}

\textbf{Proof.} Let \(P\succ0\) satisfy the program’s linear matrix inequalities and set 
\(
S_f = \langle P\rangle_E.
\)
\paragraph{Invariance with contraction.}  
For any \(x\in S_f\), \(x^\top P^{-1}x\le1\).  Then
\begin{align}
&(A_f x)^\top P^{-1}(A_f x)
= x^\top A_f^\top P^{-1}A_f x\\
& \quad \quad \quad \quad \quad \quad \quad \quad \le x^\top (\lambda_f P^{-1}) x\\
&\text{since }A_f\,P\,A_f^\top\preceq \lambda_f P= \lambda_f\,x^\top P^{-1}x\le \lambda_f\le1.
\end{align}
\paragraph{State and input inclusions.}  
For each row \(q_{\bar x,i}^\top\) of \(Q_{\bar x}\),  
\begin{align}
\max_{x\in S_f} q_{\bar x,i}^\top x
&= \max_{x^\top P^{-1}x\le1} q_{\bar x,i}^\top x\\
&= \|P^{\tfrac12}q_{\bar x,i}\|_2
\;\le\; q_{\bar x,i},
\end{align}
which is equivalent to \(Q_{\bar x}P\,Q_{\bar x}^\top\preceq\mathrm{diag}(q_{\bar x})\).  Thus \(S_f\subseteq\mathcal{S}_{\bar x}\).  Similarly,  
\begin{align}
\max_{x\in S_f} q_{\bar u,j}^\top (K_f x)
&= \max_{x^\top P^{-1}x\le1} q_{\bar u,j}^\top K_f x\\
&= \|P^{\tfrac12}K_f^\top q_{\bar u,j}\|_2
\;\le\; q_{\bar u,j},
\end{align}
so \(K_f\,S_f\subseteq\mathcal{S}_{\bar u}\).
\paragraph{Maximum volume.}  
The volume of \(\langle P\rangle_E\) is proportional to \(\det(P)^{1/2}\). With that, the proof is concluded. 
\(\hfill \Box\)

\begin{rem}\label{rem:inscribedZono}
To preserve the \emph{linear} structure of the MPC formulation, one can replace the exact ellipsoidal terminal set
\(
\mathcal S_f
\)
by a \emph{zonotopic} approximation
\(
\mathcal Z^f
\) [Theorem 2,\cite{gassmann2020scalable}].
In this way, the inclusion of the terminal set \(\;\bar x(k+N|k) \in \mathcal Z^f\;\)
can be enforced through a purely constraint \emph{linear} \( \bar x(k+N|k) - c_f \;=\; G_f\,\bm \alpha, \, -\,\mathbf{1} \le\bm \alpha\le\mathbf{1}, \) thus obtaining an LP.
\end{rem}

\begin{rem}
To avoid the dead‑beat and idle behavior often observed in LP‑based MPC~\cite{saffer2004analysis,rao2000linear}, we adopt Campo’s \(\infty/\infty\) cost formulation~\cite{campo1989model}.  A single slack variable \(\beta\) jointly bounds the cumulative state‑tracking error and actuator effort, and the optimization seeks the smallest \(\beta\), thereby balancing the two objectives adaptively. The original cost~\eqref{eq:cost} is replaced by
\begin{subequations}
\begin{align}
J^{\star} \;:=\; &\min_{\bar{\mathbf{x}},\,\bar{\mathbf{u}},\,\beta}\;
      \beta \;+\; \ell_f(k{+}N\mid k), \\[2pt]
\text{s.t.}\quad
      &\sum_{j=k}^{k+N-1} g\!\bigl(\mathcal Q,\bar x(j)\bigr)\;\le\;\beta,\\
      &\sum_{j=k}^{k+N-1} g\!\bigl(\mathcal R,\bar u(j)\bigr)\;\le\;\beta.
\end{align}
\end{subequations}
By minimizing \(\beta\), the optimizer tightens whichever cumulative norm, state or input, currently dominates, preventing persistent bias toward either zero‑input (idle) or maximum‑input (dead‑beat) behavior.  The invariance and stability results of Theorem~\ref{thm:finalSetComp} remain valid under this modified cost since $\beta$ is an upper bound on the cost $J$.
\end{rem}

Algorithm~\ref{alg:perception_mpc} addresses Problem~\ref{problem.1} by combining a linear‑programming tube‑based robust MPC with perception‑driven state estimation.  Offline, it first designs a robust observer gain \(L\) via Theorem~\ref{thm:zonotopeObserver}; then computes a robust feedback gain \(K\) (Theorem~\ref{thm:zonotopeFeedbackGain}) and a terminal gain \(K_f\) together with the maximum‑volume terminal set \(\mathcal S_f\) (Theorems~\ref{thm:finalSetComp}–\ref{thm:maxSize}).  These results allow the construction of tightened state and input sets \(\mathcal S_{\bar x}\) and \(\mathcal S_{\bar u}\) that account for the deviation set \(\mathcal S_{\tilde x}\).  Online, at each time step \(k\) the current estimate \(\hat{x}(k)\) is available; the MPC problem~\eqref{eq:MPCformulation} is solved to deliver a nominal trajectory and the control law \(u(k)\) in~\eqref{eq:MPC_control_Input}, which is applied to the plant.  Concurrently, a new measurement \(y(k)\) is obtained from a perception pipeline—image capture, feature extraction, and the map of~\eqref{eq:output}—and fused with the observer~\eqref{eq:estimator} to refresh \(\hat{x}(k)\).  The algorithm outputs the closed‑loop state path \(\{x(k)\}_{k=0}^{N_s}\), the applied input sequence \(\{u(k)\}_{k=0}^{N_s-1}\), and the cumulative cost \(J^{s}\) defined in~\eqref{eq:reportCost}.

\begin{algorithm2e}[h]
\caption{Perception-MPC Algorithm}
\label{alg:perception_mpc}
\textbf{Input:} Initial state estimate \(\hat{x}_0\),  prediction horizon \(N\), system matrices \(A\), \(B\), \(C\); constraint sets $(\mathcal{S}_x,\mathcal{S}_u, \mathcal{S}_e, \mathcal{S}_{\tilde x})$; cost sets $(\mathcal{Q},\mathcal{R}, \mathcal{P})$; process noise zonotope \(\mathcal{Z}^w_c\) and measurement noise zonotope \(\mathcal{Z}^v_c\).
\\ \textbf{Offline Computations:}
    \begin{enumerate}
        \item Compute the robust observer gain \(L\) by solving Theorem~\ref{thm:zonotopeObserver}.
        \item Compute the robust feedback gain \(K\) by solving Theorem~\ref{thm:zonotopeFeedbackGain}.
        \item Calculate the tightened input constraint set \(\mathcal{S}_{\bar u}\) and state constraint set \(\mathcal{S}_{\bar x}\).
        \item Compute the terminal gain \(K_f\) by solving Theorem~\ref{thm:finalSetComp}.
        \item Find the maximum size terminal set $\mathcal{S}_f$ using Theorem~\ref{thm:maxSize} and Remark~\ref{rem:inscribedZono}.
    \end{enumerate}

\textbf{Online MPC Loop:}

\For{\(k = 0,1,\dots,N_s-1\)}
    {Obtain the current state estimate \(\hat{x}(k)\),\\
    {Solve MPC Optimization in \eqref{eq:MPCformulation}},    \\ 
    \textbf{Plant Update:} Apply \(u(k)\) from \eqref{eq:MPC_control_Input} to the plant,\\
    \textbf{Measurement and Estimation:} 
        \begin{enumerate}
        \item Obtain picture based on \eqref{eq:perceptionBased}
        \item Pass through the perception map \eqref{eq:output} 
        
        and obtain $y(k)$   
            \item Update the state estimate using the observer \eqref{eq:estimator}
        \end{enumerate}}
 \textbf{Output:} The state trajectory \(\{x(k)\}_{k=0}^{N_s}\), the control inputs \(\{u(k)\}_{k=0}^{N_s-1}\), and the cumulative cost $J^s$ as in \eqref{eq:reportCost}.
\end{algorithm2e}

\section{Experimental Results}
This section first assesses the state estimator derived from Theorem~\ref{thm:zonotopeObserver} and compares its performance against a standard Kalman filter estimator that presumes Gaussian sensor noise.  We then demonstrate the MPC–Perception framework of Algorithm~\ref{alg:perception_mpc} in a realistic setting: plant states are inferred from recorded camera images via a neural‑network‑based perception observer, and the resulting control inputs drive the closed loop.  A comprehensive comparison with an MPC scheme employing Kalman‑based state estimates is provided, including cumulative costs and both estimated and ground‑truth trajectories.

\subsection{Observer Design Simulation Example}

\begin{figure}[t]
        \centering
        \includegraphics[width=\linewidth]{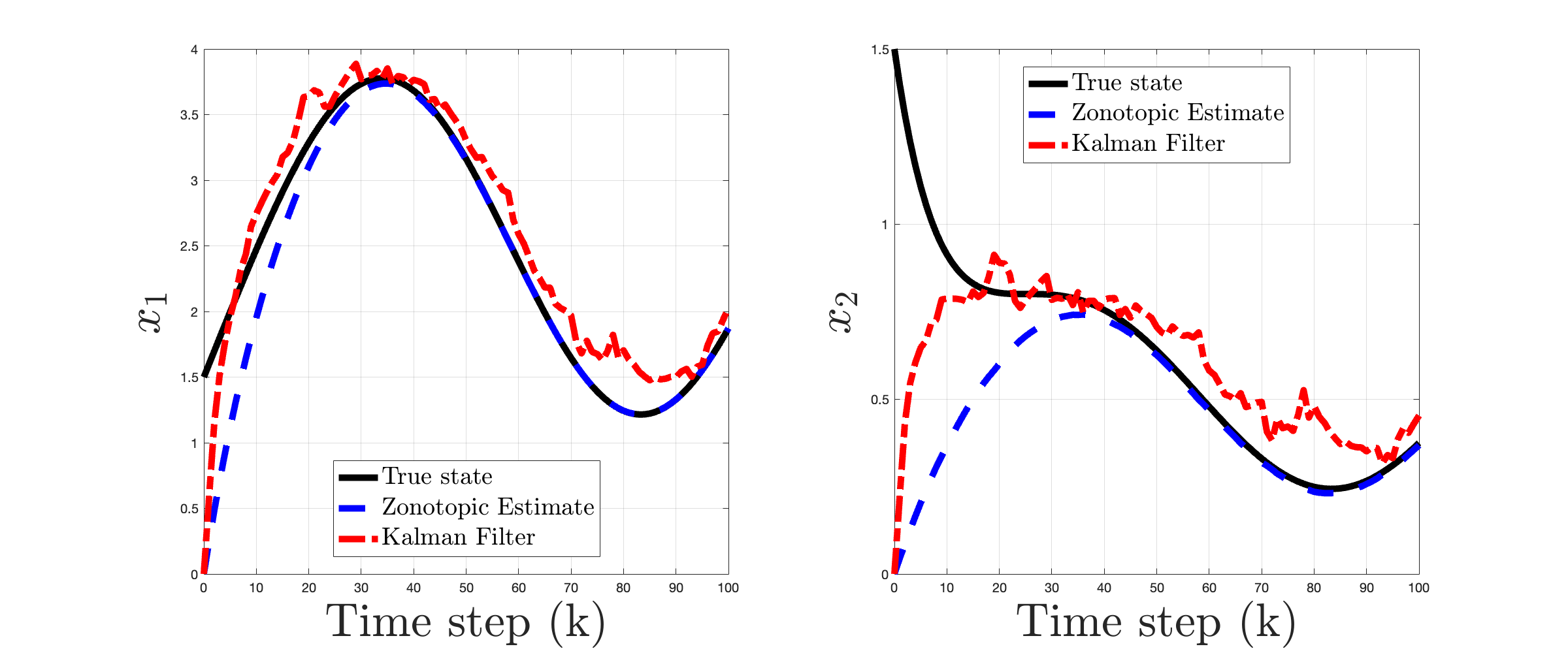}
        \caption{State estimation comparison between the zonotopic observer and Kalman gain.}
        \label{fig:state_estimation}
\end{figure}

Consider a linear system as 
\(    A = \begin{bmatrix} 0.9 & 0 \\ 0 & 0.9 \end{bmatrix}, \, 
    B = \begin{bmatrix} 0.5 \\ 0.1 \end{bmatrix}, \,  
    C = \begin{bmatrix} 1 & 0.5 \end{bmatrix}, \, w(k) = 0. 
\)
To highlight the contrast between our zonotopic estimation algorithm and the conventional Gaussian‑noise assumption. Specifically, we design the observer according to Theorem~\ref{thm:zonotopeObserver}.
Unlike conventional Gaussian noise assumptions, we model perception-based sensor noise using a non-Gaussian distribution  
\(
    v(k) = c_v + z(k), \quad z(k) \sim \text{Laplace}(0, b), \quad b = \frac{G_v}{\sqrt{2}},  
\)
where \( c_v = 0.5 \) represents a deterministic bias, and \( G_v = 0.5 \) defines the sensor noise spread. The Laplace distribution introduces heavy-tailed characteristics, making state estimation more challenging. The initial estimation error is modeled as a zonotope following Assumption~\ref{ass:ErrorBound}  
\(
    x_0 = \begin{bmatrix} 1.5 & 1.5 \end{bmatrix}^\top, \quad  
    \hat{x}_0 = \begin{bmatrix} 0 & 0 \end{bmatrix}^\top, \quad  
    c_e = x_0 - \hat{x}_0, \quad  
    G_e = 3.5 I_2.  
\)
The sensor noise is also represented as a zonotope following Assumption~\ref{ass:zonotopicnoise}, with parameters  
\(
    c_v = 0.5, \quad G_v = 0.5, \quad F_v = 0, \quad \theta_e = 0.  
\)
Two estimation methods are considered: 1. Zonotopic Observer – Implemented using Theorem~\ref{thm:zonotopeObserver} with contraction parameter \( \lambda_L = 0.95 \). 2. Kalman gain – Assumes Gaussian noise and determines the observer gain via the discrete-time Riccati equation where \( Q = 10^{-2} I \) and \( R = G_v^2 \). The system is simulated for \( N_{\text{sim}} = 100 \) steps with a time-varying control input 
\(
    u(k) = 0.5 + 0.3 \sin \left( \frac{2\pi k}{N_{\text{sim}}} \right).  
\)
At each time step \(k\), the true state \(x(k)\), the zonotopic estimate \(\hat{x}(k)\), and the Kalman estimate \(\hat{x}^K(k)\) are computed. Figure~\ref{fig:state_estimation} illustrates the performance of the proposed zonotopic observer in Theorem~\ref{thm:zonotopeObserver}. The zonotopic observer provides bounded estimates under heavy-tailed noise, whereas the Kalman filter degrades under incorrect Gaussian noise assumptions. The zonotopic observer converges more slowly to the reference trajectory because of its conservative gain compared to the Kalman gain.

\subsection{Real-world Implementation}
We validate our theoretical results through control-from-pixels experiments, where the perception module generates output readings as defined in \eqref{eq:output}. The following sections detail the experimental setup and present the results.
\subsubsection{Hardware and Setup}
Algorithm~\ref{alg:perception_mpc} is implemented on a Husarion ROSbot XL with Jetson Orin Nano (NVIDIA, California, USA), an omnidirectional robot with a fully known kinematics model, using a ROS2 framework. An Astra Pro Plus camera (Orbbec Inc., Shenzhen, China) on the master computer captures perception data, and commands are communicated to the robot via ROS2 over Wi-Fi.

\begin{figure}[t!]
    \centering
        \centering
        \subfloat[]{{\includegraphics[height=2.4in]{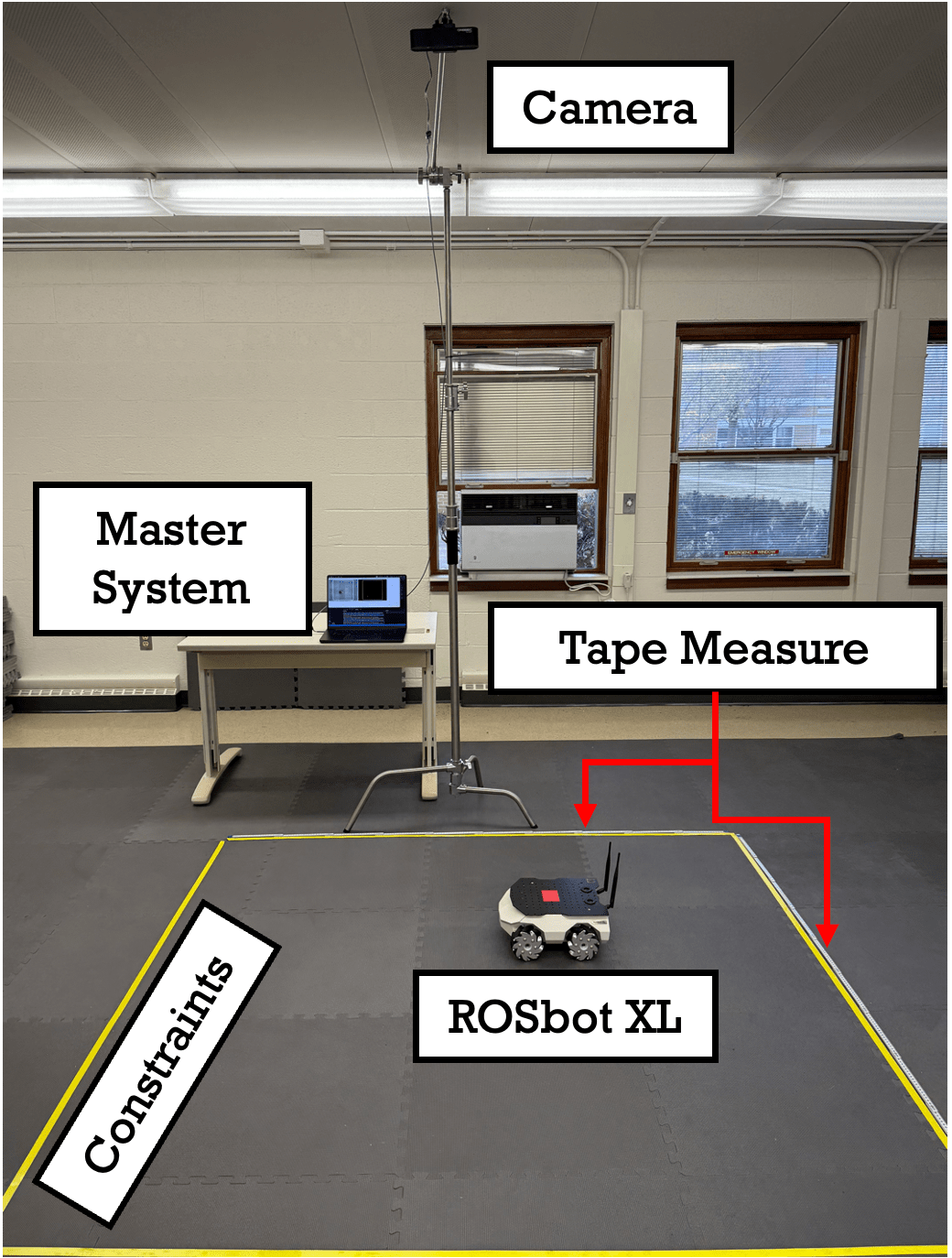} }}%
        \subfloat[]{{\includegraphics[height=1.1in]{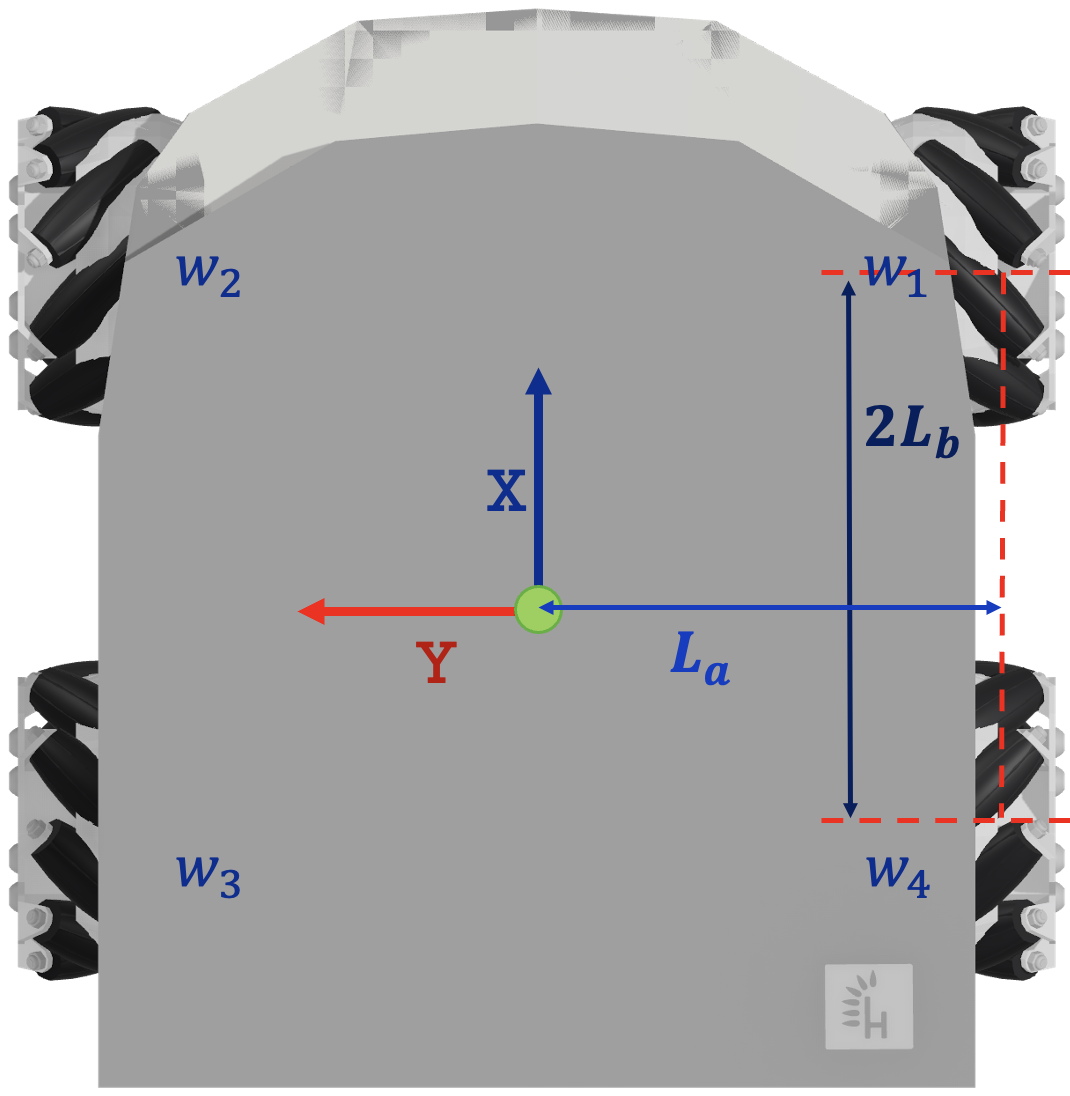} }}%
    \caption{(a) Experimental setup for the Perception-MPC system; the ROSbot~XL operates in a bounded area with an overhead camera for state estimation and a master system for control, communicating using ROS2, and (b) Top view of the Husarion ROSbot XL, featuring its coordinate frame, labeled wheels, and body length.}
    \label{fig:PlantAndRobot}
\end{figure}
\noindent The discrete-time state space kinematics of the omnidirectional robot is \cite{killpack2010visual}
\(
    A = I_{3\times 3}, B = \begin{bmatrix}
        t_s L_{ab}& t_s L_{ab}& t_s L_{ab}& t_s L_{ab}\\
        t_s L_{ab}& -t_s L_{ab}& t_s L_{ab}& -t_s L_{ab}\\
        t_s & -t_s & -t_s & t_s
    \end{bmatrix},
\)
where $L_{ab}:= L_a + L_b$, with $L_a = 135mm$ and $L_b = 85mm$. The states are $\begin{bmatrix}
    x&y& \theta
\end{bmatrix}^{\top}$, $C = I_{3\times 3}$, and $t_s=0.35 s$ is the time step for discretization. For this experiment, the first two states, \(x\) and \(y\), are obtained via the perception map, while the third state, \(\theta\), is measured using a SLAM-based algorithm by separating orientation and position \cite{agarwal2017rfm}. We assume we have an exact reading of \(\theta\) at each step. More specifically, this subsection evaluates the observer's performance when incorporating a perception-based module to determine the position of the robot. The robot's initial state is given by 
\(
x(0)= \begin{bmatrix} 0.75 & 0.75 & 0 \end{bmatrix}^\top,
\)
and we assume that the initial state estimate is identical to the true state, i.e., 
\(
\hat{x}(0) = x(0).
\) The prediction horizon is \(N = 12\) time steps, and the perception module captures and processes images at $30\text{Hz}$.
\subsubsection{Perception Data and Convolutional Neural Network Map}
\begin{figure}[t]
        \centering
        \includegraphics[width=0.7\linewidth]{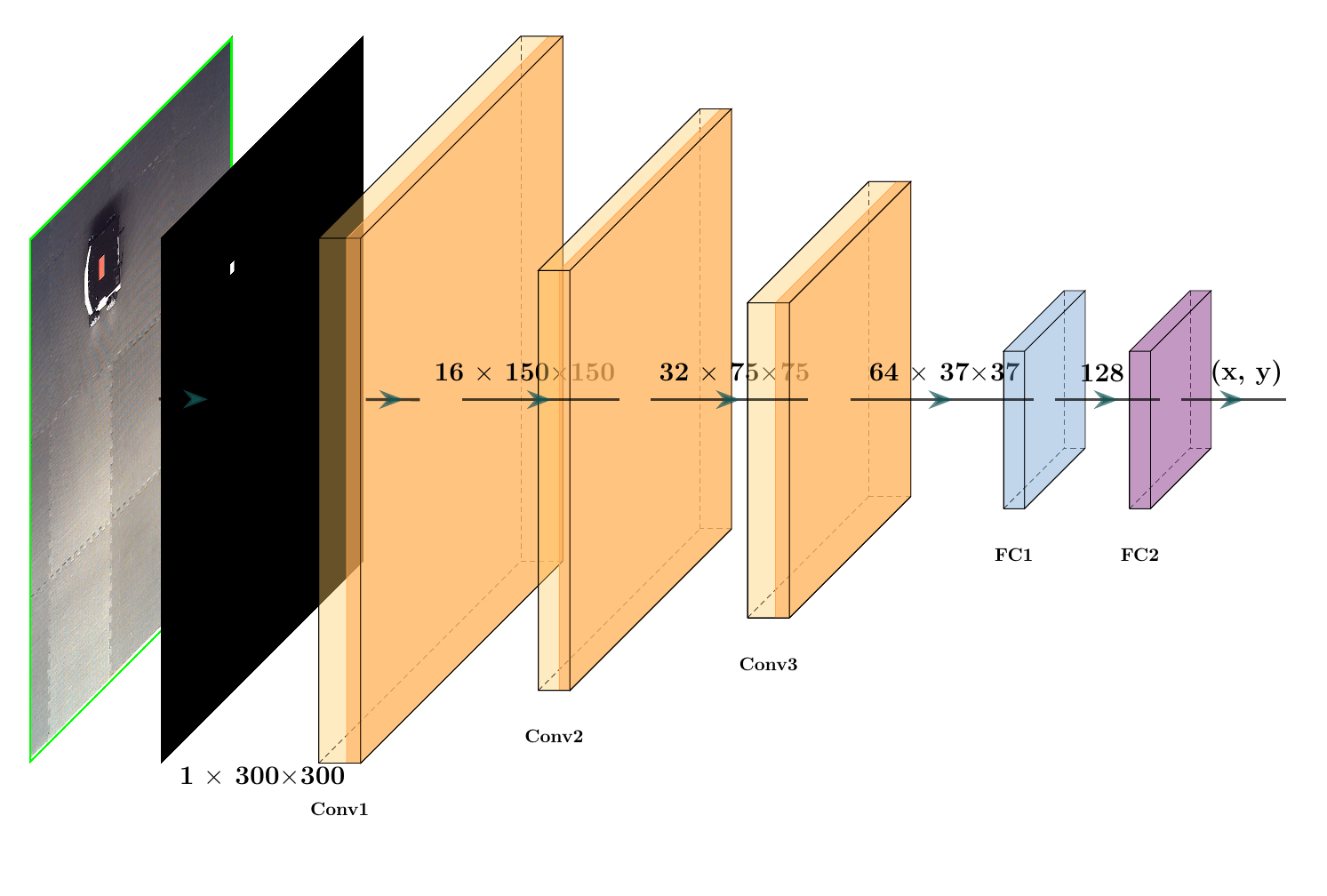}
        \caption{\texttt{RobotPerceptionNet} architecture.}
        \label{fig:network}
\end{figure}

A dataset is collected and used to train \texttt{RobotPerceptionNet} using PyTorch \cite{paszke2019pytorch} that regresses the physical state (position) of a Husarion ROSbot XL. The dataset and network are designed to bridge the gap between physical coordinates in a 2\(\times\)2 $m^2$ space and their corresponding 300\(\times\)300 pixel image representations. The collected images are preprocessed so that the red sticker on the robot is transformed into an 8\(\times\)8 pixel white area on a black background. The data collection process is detailed in the following sections, a summary of the dataset parameters is provided, and the neural network's architecture is described. The dataset is obtained from a real-world system using the ROSbot XL within a bounded area, as shown in Fig.~\ref{fig:PlantAndRobot}. The robot is positioned at various locations, with its coordinates recorded within \(x_m \in [-1, 1]\) and \(y_m \in [-1, 1]\) along with the corresponding images. A transformation is applied before training so that the robot's location is represented as an 8\(\times\)8 white block on a black background, thereby enhancing accuracy. Finally, the dataset is partitioned into training and testing sets, with the parameters summarized in Table~\ref{tab:dataset}.

\begin{table}[H]
\centering
\begin{tabular}{lcc}
\hline
\textbf{Dataset} & \textbf{Number of Samples} & \textbf{Image Dimensions} \\
\hline
Training & 3000 & 300~$\times$~300 \\
Testing  & 200  & 300~$\times$~300 \\
\bottomrule
\end{tabular}
\vspace{2pt}
\caption{Dataset summary with training and testing splits.}
\label{tab:dataset}
\end{table}

\noindent The perception map, \texttt{RobotPerceptionNet}, is designed to directly predict the robot's physical coordinates from images with an emphasis on speed of estimation. Inspired by existing literature \cite{hijazi2015using}, the network is structured with two main components: convolutional layers for feature extraction and fully connected layers for regression. Fig.~\ref{fig:network} and Table~\ref{tab:cnn_arch} summarize the designed network architecture. The network is trained for 25 epochs with a batch size of 64, using the Adam optimizer (\(\mathrm{lr}=10^{-3}\)) to minimize the mean‑squared‑error (MSE) loss.

\begin{table}[t]
  \centering
  \caption{Convolutional neural network used for image–to–state mapping.}
  \label{tab:cnn_arch}
  \renewcommand{\arraystretch}{1.05}
  \begin{tabular}{@{}lccc@{}}
    \toprule
    \textbf{Layer} & \textbf{Output shape} & \textbf{Kernel / Pool} & \textbf{Activation} \\ \midrule
    Input          & $1 \times 300 \times 300$ & --               & --   \\
    Conv\,1        & $16 \times 300 \times 300$ & $3\!\times\!3$, pad 1 & ReLU \\
    MaxPool\,1     & $16 \times 150 \times 150$ & $2\!\times\!2$        & --   \\
    Conv\,2        & $32 \times 150 \times 150$ & $3\!\times\!3$, pad 1 & ReLU \\
    MaxPool\,2     & $32 \times 75  \times 75$  & $2\!\times\!2$        & --   \\
    Conv\,3        & $64 \times 75  \times 75$  & $3\!\times\!3$, pad 1 & ReLU \\
    MaxPool\,3     & $64 \times 37  \times 37$  & $2\!\times\!2$        & --   \\
    Flatten        & $87\,616$                  & --               & --   \\
    FC\,1          & $128$                     & --               & ReLU \\
    FC\,2          & $2$                       & --               & --   \\ \bottomrule
  \end{tabular}
\end{table}

\begin{figure}[t!]
    \centering
    \subfloat[]{\includegraphics[width=0.5\columnwidth]{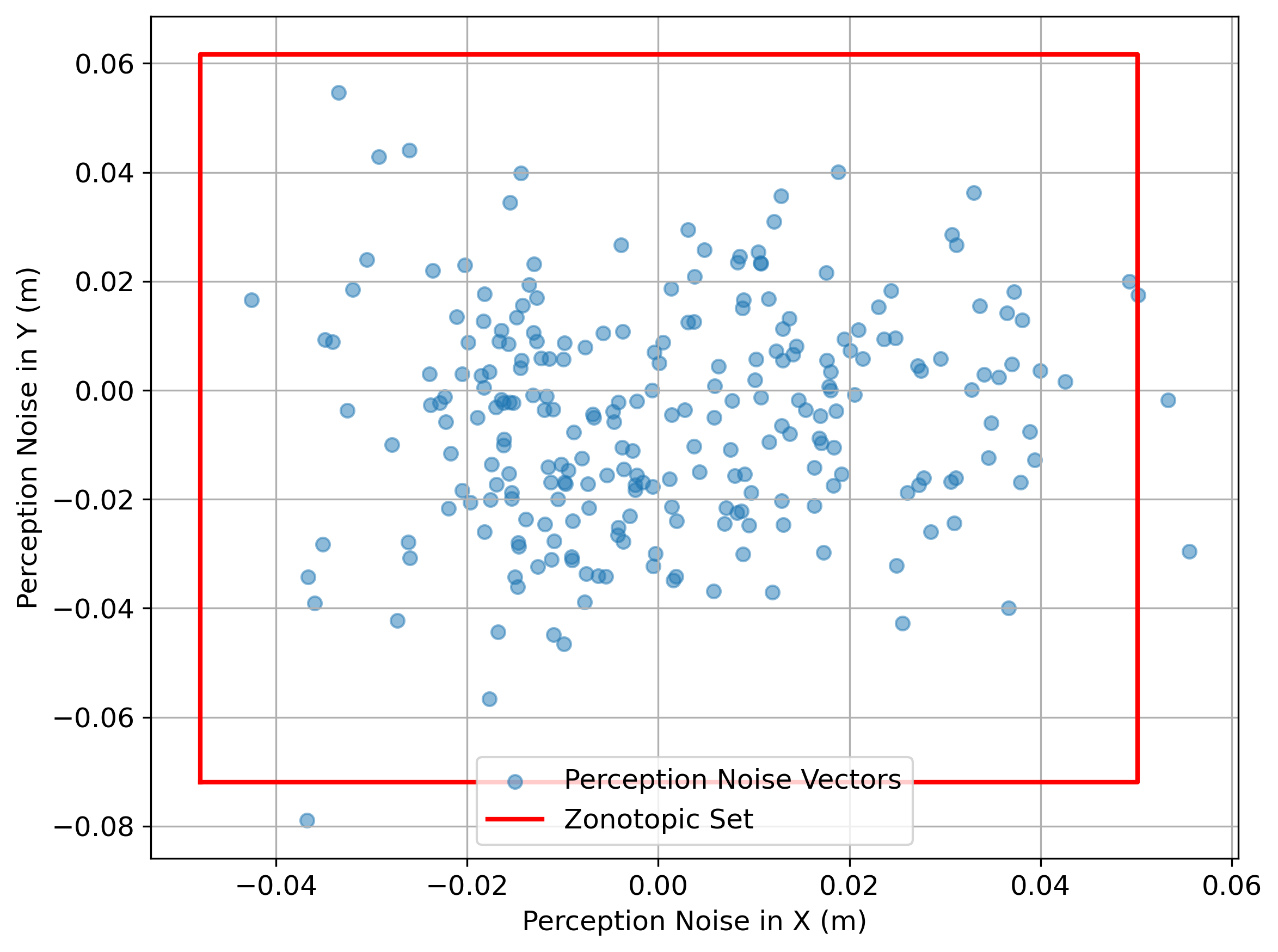}}
    \subfloat[]{\includegraphics[width=0.5\columnwidth]{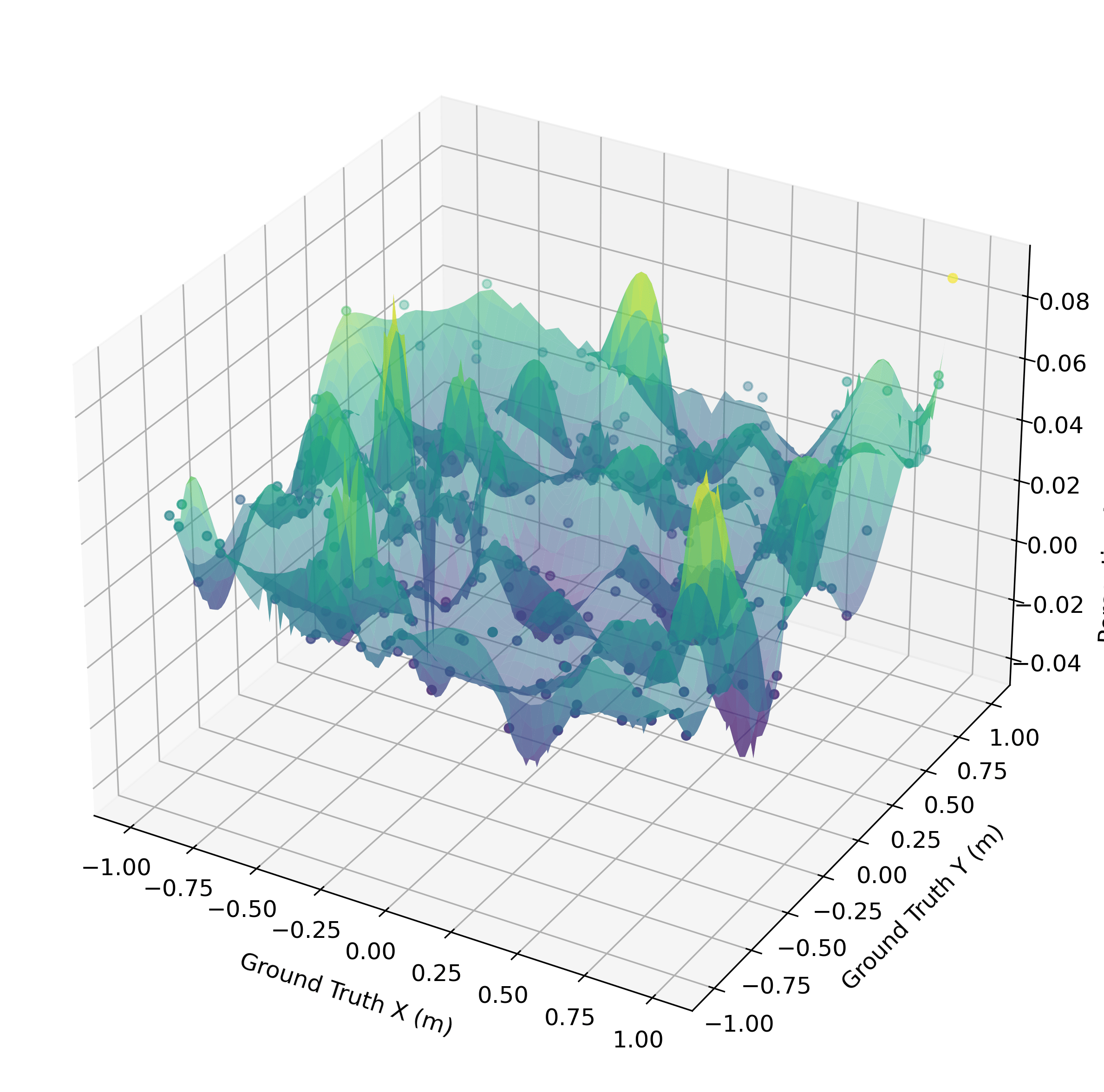}}\par
    \vspace{-5pt}
    \subfloat[]{\includegraphics[width=0.5\columnwidth]{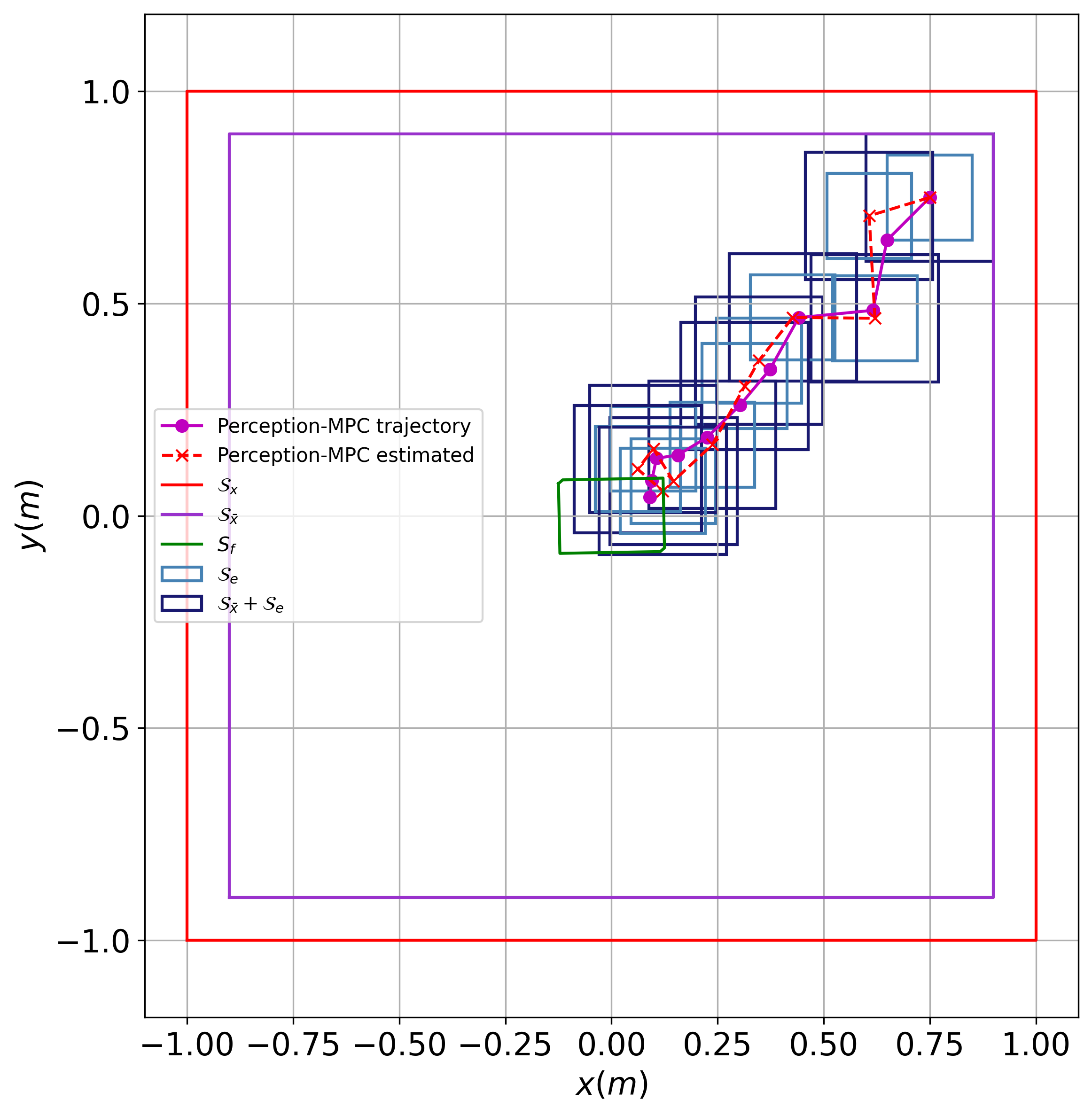}}
    \subfloat[]{\includegraphics[width=0.5\columnwidth]{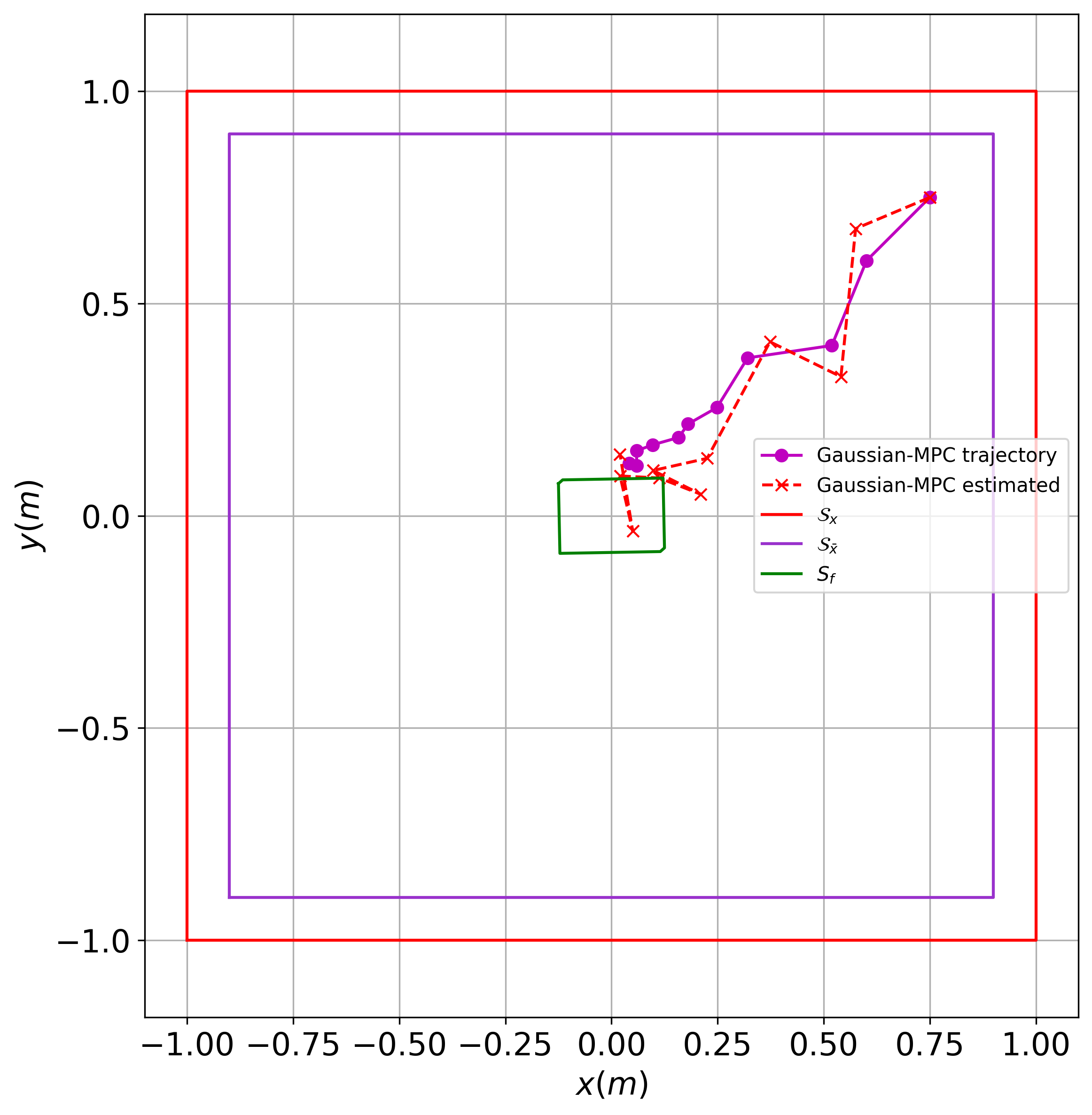}}
    \caption{
    (a) Performance of \texttt{RobotPerceptionNet} on the test dataset, showcasing the perception noise and its corresponding zonotopic representation. (b) A 3D plot illustrating the variation of perception noise magnitude relative to the ground truth locations. A comparative analysis of the trajectories from (c) Perception-MPC and (d) Gaussian-MPC, depicting both the estimated and actual states.
    }
    \label{fig:TrainingResults}
\end{figure}

\noindent Upon completing training, we evaluate the model on the held‑out test set (Table~\ref{tab:dataset}). We then compute the Euclidean perception error in world coordinates and visualized its distribution with a 2D zonotopic plot (Fig.~\ref{fig:TrainingResults}(a)) and a 3D surface rendering (Fig.~\ref{fig:TrainingResults}(b)). The zonotopic set is described by
\begin{align}\label{eq:zonotopeSensorNoise_exp}
    \mathcal{Z}^v = \langle \begin{bmatrix}
        0.0011 & -0.0051 & 0 
    \end{bmatrix}, \begin{bmatrix}
        0.0490 & 0     & 0\\
        0      & 0.0667& 0\\
        0      & 0     & 0
    \end{bmatrix} \rangle_Z,
\end{align}
which is computed using $95\%$ of the perception noise data points.


\subsubsection{Computing the Invariant Sets and Associated Gains}
After estimating the process noise set (Assumption~\ref{ass:zonotopicProcessnoise}) and the perception measurement noise (Assumption~\ref{ass:zonotopicnoise}), we determine an observer gain \(L\) and compensation feedback term \(K\) so that the estimation error set \(\mathcal{S}_e\) and state deviation set \(\mathcal{S}_{\tilde{x}}\) remain bounded. This is achieved using Theorems~\ref{thm:zonotopeObserver} and \ref{thm:zonotopeFeedbackGain}. For this purpose, we consider $\mathcal{Z}^v$ as defined in \eqref{eq:zonotopeSensorNoise_exp}. Since the robot experiences no disturbances, we set $w(k)=0$ for all $k \geq 0$. The constraints and initial sets are defined as follows.

1) State constraint: $\mathcal{Z}^x = \langle [0, 0, 0], \operatorname{diag}(1, 1, 1) \rangle_Z$.

2) Input constraint: $\mathcal{Z}^u = \langle [0, 0, 0, 0], \operatorname{diag}(1, 1, 1, 1) \rangle_Z$.

3) Initial estimation set: $\mathcal{Z}^e = \langle [0, 0, 0], \operatorname{diag}(0.1, 0.1, 0) \rangle_Z$.

4) Initial state deviation set: $\mathcal{Z}^{\tilde{x}} = \langle [0, 0, 0], \operatorname{diag}(0.2, 0.2, 0) \rangle_Z$.

\noindent By selecting $\lambda_L = 0.85$, Theorem~\ref{thm:zonotopeObserver} yields \[ L = 
\begin{bmatrix}
0.9824 & 0.0003 & 0.0000 \\
0.0001 & 0.9879 & 0.0000 \\
0.0000 & 0.0000 & 0.0000
\end{bmatrix}.
\] By selecting $\lambda_{\tilde{x}} = 0.9$ Theorem~\ref{thm:zonotopeFeedbackGain} yields
\[
K = \begin{bmatrix}
-3.2468 & -3.2468 & 0.0000 \\
-3.2468 & 3.2468  & 0.0000 \\
-3.2468 & -3.2468 & 0.0000 \\
-3.2468 & 3.2468  & 0.0000
\end{bmatrix}.
\]  
We choose the following zonotopic sets for the stage and terminal cost sets as
$\mathcal{P} = \langle [0\; 0\; 0], 1e{-4}\times I \rangle_Z, \mathcal{Q} = \langle [0\; 0\; 0], 2e1\times I \rangle_Z, \mathcal{R} = \langle [0\; 0\; 0\; 0], 5e1\times I \rangle_Z$. Theorem~\ref{thm:finalSetComp} yields
\[
K_f = \begin{bmatrix}
 -0.5931 & -0.8311 & -0.4837\\
 -0.6539 & 0.7942 & 0.8790\\
 -0.5931 &-0.8311 & 0.4837\\
 -0.6539 & 0.7942 & -0.8790
\end{bmatrix}.
\]
 By selecting $\lambda_{f} = 0.95$ and the terminal set $\mathcal{S}_f$ is computed as a maximum size ellipsoidal set $\langle P \rangle_E$ from Theorem~\ref{thm:maxSize} with 
 \[P = 
\begin{bmatrix}
0.0420 & 0.0004 & -0.0011 \\
0.0004 & 0.0203 & 0.0001 \\
-0.0011 & 0.0001 & 0.0235
\end{bmatrix}
\],
and the associated inscribed zonotopic set is found using $3$ generators via Remark~\ref{rem:inscribedZono} to keep the MPC an LP.

\subsubsection{MPC-Perception Performance Comparison} To demonstrate the effectiveness and advantages of our proposed methodology over traditional probabilistic approaches—which rely on the assumption of Gaussian-distributed measurement and process noise (e.g., \cite{yan2005incorporating}), we conduct a comparative study against an approach, referred to as Gaussian-MPC. In the Gaussian-MPC framework, the gains \(L\), \(K\), and \(K_f\) are computed using LQR with cost weights set as \(Q_L = 10^2 I\), \(R_L = 0.1 I\) for \(L\), and \(Q_K = 10^2 I\), \(R_K = 0.1 I\) for both \(K\) and \(K_f\).

\noindent Fig.~\ref{fig:TrainingResults}(c) and (d) present a comparison between the Perception-MPC and Gaussian-MPC approaches. As illustrated, the estimated states under Perception-MPC remain within the set \(\mathcal{S}_e \oplus \mathcal{S}_{\tilde{x}}\), whereas Gaussian-MPC occasionally exhibits larger estimation errors. This difference is reflected in the overall performance, as summarized in Table~\ref{table:performances}, where the cost is evaluated using
\begin{align}\label{eq:reportCost}
    J^s &= \sum_{i=0}^{N_s-1} \ell^s(i) + \ell^s_f,
\end{align}
with stage cost \(\ell^s(i) := g(\mathcal{Q}, x(i)) + g(\mathcal{R}, u(i))\), terminal cost \(\ell^s_f := g(\mathcal{P}, x(N_s))\), and \(N_s\) denoting the number of time steps in the experiment.

\begin{table}[ht]
\centering
\resizebox{0.3\textwidth}{!}{%
\begin{tabular}{c c c}
\toprule
\textbf{MPC Type}  &  \quad \quad \quad \quad & \textbf{Cost $(J^s)$} \\ 
\midrule
Perception-MPC     &  & 9019.5945         \\
Gaussian-MPC       &  & 12320.2557       \\ 
\bottomrule
\end{tabular}%
}
\vspace{2pt}
\caption{Performance costs $(J^s)$ for Perception-MPC and Gaussian-MPC.}
\label{table:performances}
\end{table}

\begin{figure}[t]
        \centering
        \includegraphics[width=0.8\linewidth]{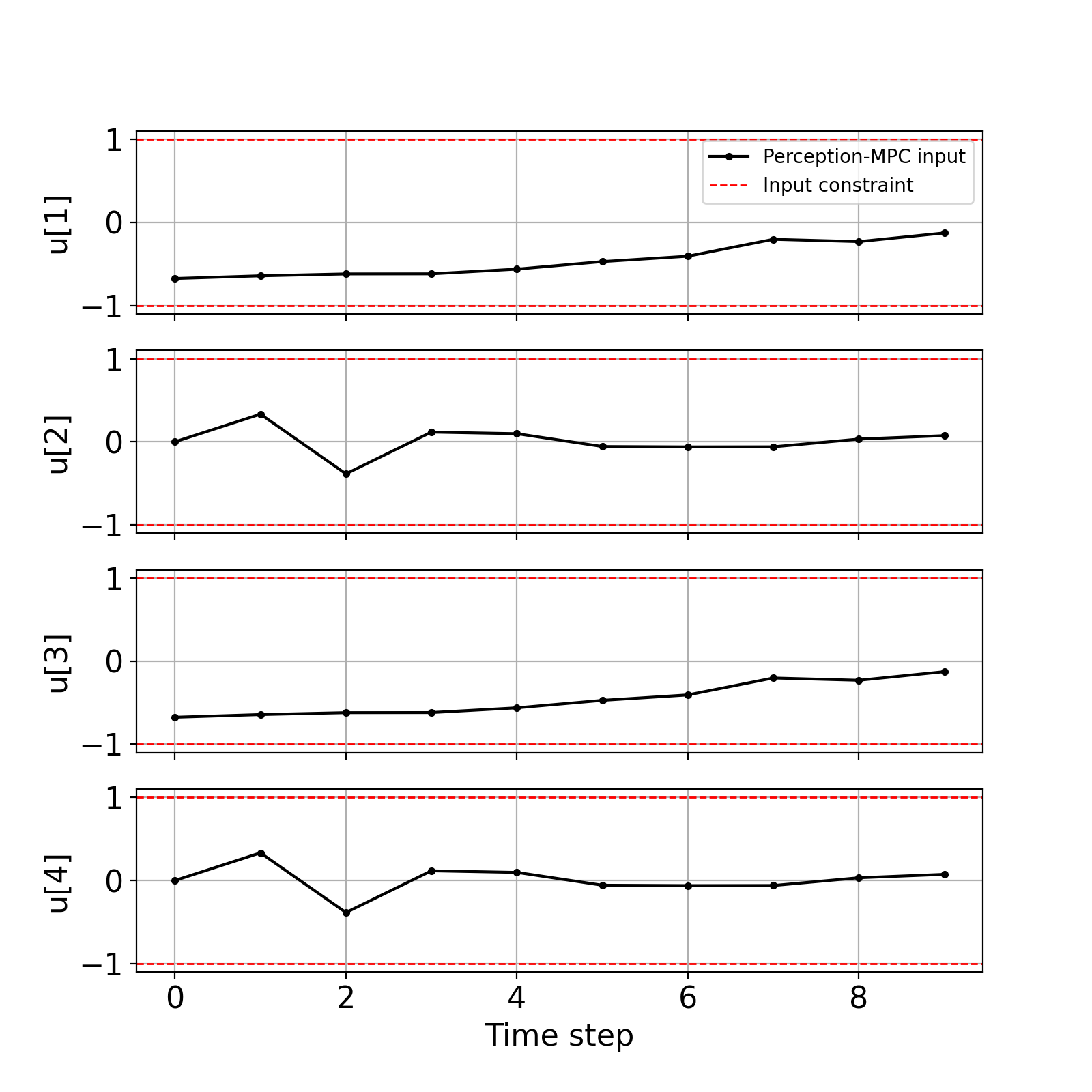}
        \caption{Control inputs produced by Perception-MPC alongside the imposed input constraints.}
        \label{fig:control_input}
\end{figure}

\noindent Additionally, Fig.~\ref{fig:control_input} confirms that the control inputs generated by Perception-MPC remain within the prescribed constraints throughout the task.

\begin{rem}
    Assuming Gaussian noise results in an unbounded support. Stochastic tube MPC variants, for example, \cite{cannon2012stochastic}, address this by enforcing chance constraints, yet the non-zero mean perception noise in \eqref{eq:zonotopeSensorNoise_exp} introduces estimator bias that such schemes cannot remove or consider. In addition, system behavior is highly sensitive to the state and input covariance weights \(Q\) and \(R\); some choices work, others render the problem infeasible.  
\end{rem}

\section{Conclusion}
In this paper, we present a novel control framework that integrates CNN-based perception for state estimation with a robust zonotopic observer and an LP-based MPC strategy. Our approach systematically addresses the uncertainty and robustness challenges inherent in systems employing perception-based observers. While existing methods treat perception outputs as signals corrupted by zero-mean Gaussian noise, our theoretical analysis and real-world experiments demonstrate that set-based approaches yield more reliable performance. In particular, the use of constrained zonotopes to represent uncertainty and invariant sets allows for a more accurate and computationally efficient formulation of the control problem, ensuring that the true state remains within prescribed bounds despite significant disturbances.

Extending the framework to nonlinear systems and applying it to complex platforms with adaptive process noise level estimation are promising directions for future research. Further experimental validation on diverse robotic platforms and under varied conditions could also enhance the assessment of its robustness and scalability.

\bibliographystyle{IEEEtran}

\bibliography{Refs}

\begin{IEEEbiography}[{\includegraphics[width=1in,height=1.25in,clip,keepaspectratio]{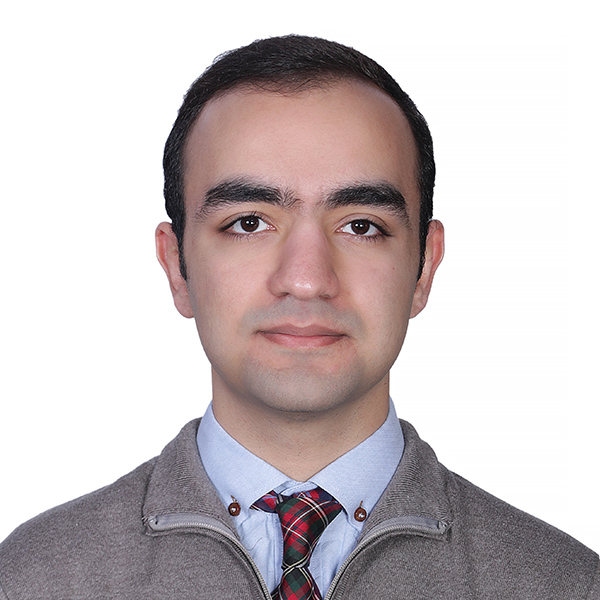}}]{Nariman Niknejad} received his bachelor's degree in Mechanical Engineering from K.N. Toosi University of Technology in Tehran, Iran, in 2020. He then completed his master's in Biosystems Engineering at Auburn University, Auburn, AL, in 2022. Currently, he is pursuing a Ph.D. in Mechanical Engineering at Michigan State University, East Lansing, MI, USA. His research centers on data-driven control, deep learning, reinforcement learning, and motion planning. Nariman has also contributed as a reviewer for several journals, including IEEE Transactions on Control Systems Technology, IEEE Transactions on Systems, Man, and Cybernetics, IEEE Control Systems Letters, and Transactions on Machine Learning Research.

\end{IEEEbiography}

\begin{IEEEbiography}[{\includegraphics[width=1in,height=1.25in,clip,keepaspectratio]{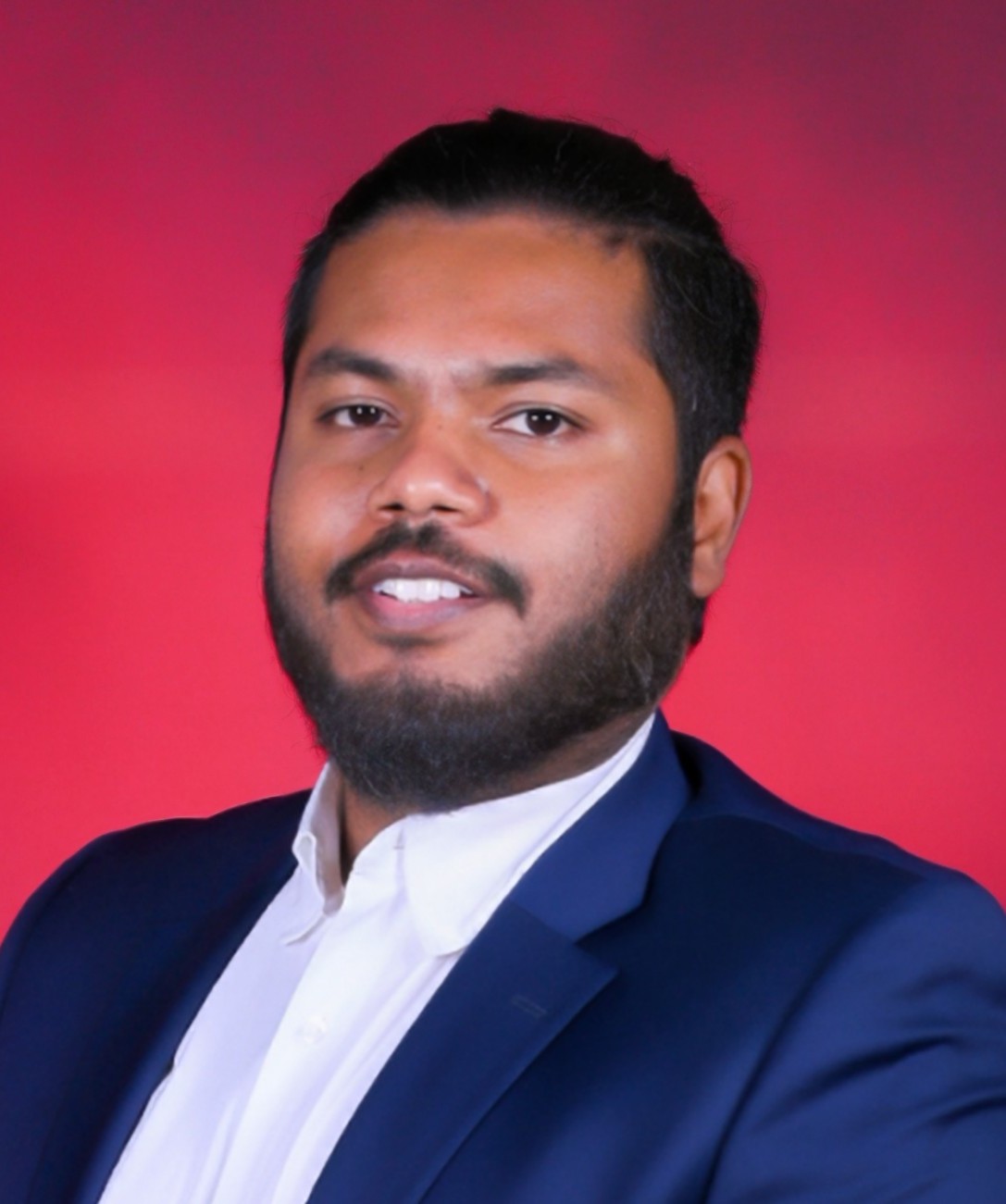}}]{Gokul S. Sankar} is a Motion Planning Engineer at Ford Motor Company, where he works at the intersection of cutting-edge autonomy and intelligent control. He earned his Bachelor's degree in Electronics and Instrumentation Engineering from Anna University, Chennai, India, in 2010, followed by an M.S. from the Indian Institute of Technology Madras in 2013, and a Ph.D. from the University of Melbourne, Australia, in 2019.
 
From 2020 to 2022, Gokul was a Research Engineer at Traxen Inc., USA, where he played a key role in developing advanced planning and control architectures for autonomous long-haul semi-trucks. Prior to that, he was a Research Fellow at the University of Michigan, Ann Arbor, from 2018 to 2019. His interests span reinforcement learning, optimal control, and robust control, with a strong focus on autonomous vehicles and cyber-physical systems.

\end{IEEEbiography}

\begin{IEEEbiography}[{\includegraphics[width=1in,height=1.25in,clip,keepaspectratio]{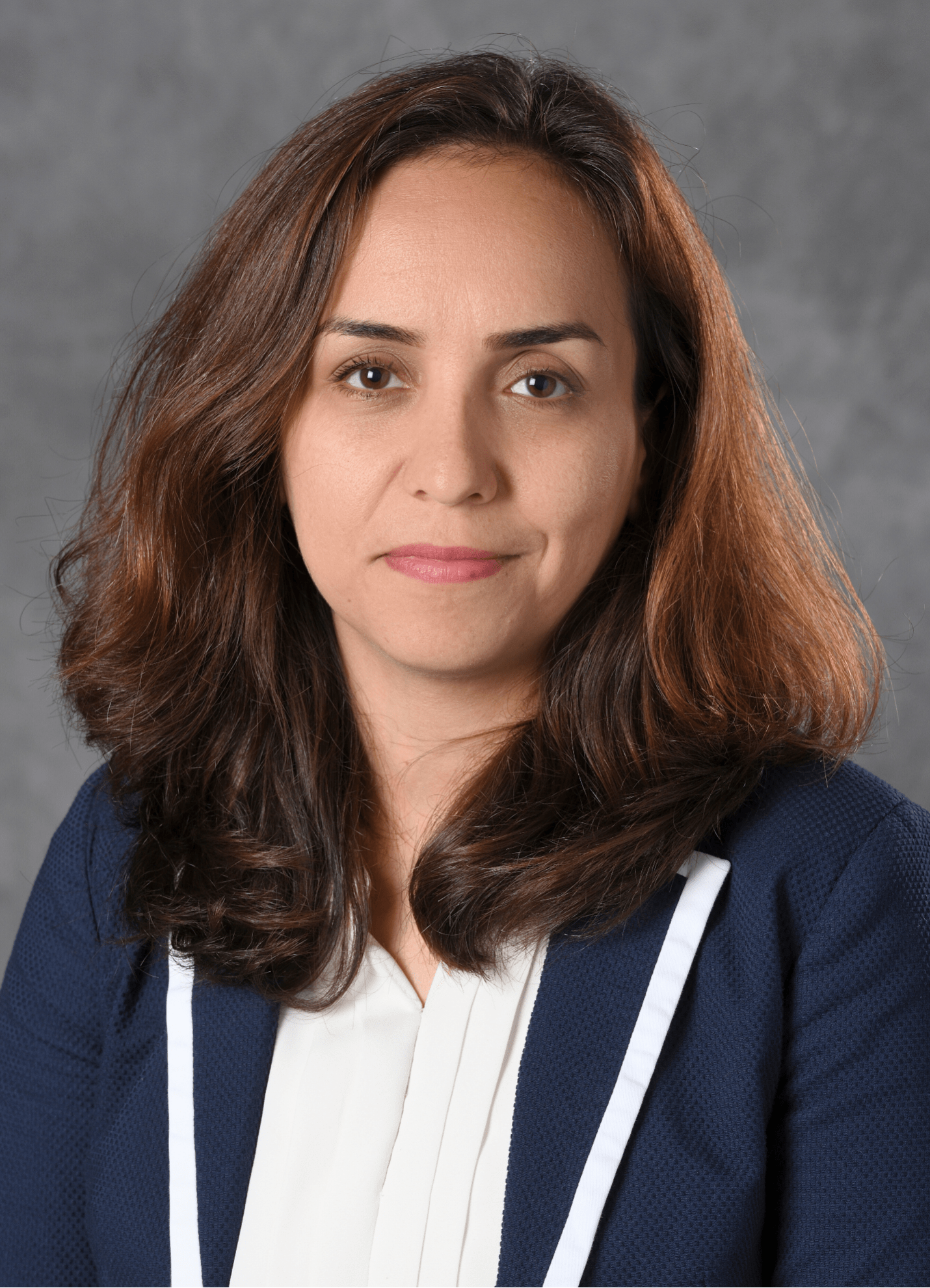}}]{Bahareh Kiumarsi} 
received her B.S. degree in Electrical Engineering from Shahrood University of Technology, Shahrood, Iran, in 2009, her M.S. degree from Ferdowsi University of Mashhad, Iran, in 2013, and her Ph.D. degree from the University of Texas at Arlington, Arlington, TX, USA, in 2017. She is currently an Assistant Professor in the Department of Electrical and Computer Engineering at Michigan State University. Prior to joining Michigan State, she was a Postdoctoral Research Associate at the University of Illinois at Urbana-Champaign. Her research interests include learning-based control, the security of cyber-physical systems, and distributed control of multi-agent systems. She serves as an Associate Editor for Neurocomputing.
\end{IEEEbiography}

\begin{IEEEbiography}[{\includegraphics[width=1in,height=1.25in,clip,keepaspectratio]{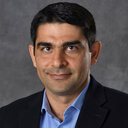}}]{Hamidreza Modares}
 received a B.S. degree from the University of Tehran, Tehran, Iran, in 2004, an M.S. degree from the Shahrood University of Technology, Shahrood, Iran, in 2006, and a Ph.D. degree from the University of Texas at Arlington, Arlington, TX, USA, in 2015, all in Electrical Engineering. He is currently an Associate Professor in the Department of Mechanical Engineering at Michigan State University. Before joining Michigan State University, he was an Assistant professor in the Department of Electrical Engineering at Missouri University of Science and Technology. His current research interests include reinforcement learning, safe control, machine learning in control, distributed control of multi-agent systems, and robotics. During the past five years, he has served as an Associate Editor for IEEE Transactions on Neural Networks and Learning Systems, Neurocomputing, and IEEE Transactions on Systems, Man, and Cybernetics: Systems.
\end{IEEEbiography}

\end{document}